\documentclass[sn-basic]{sn-jnl}

\usepackage{graphicx} 
\usepackage{xcolor}
\usepackage{makecell}
\usepackage{lscape}
\usepackage{hyperref}
\usepackage{multirow}%
\usepackage{amsmath,amssymb,amsfonts}%
\usepackage{amsthm}%
\usepackage{mathrsfs}%
\usepackage[title]{appendix}%
\usepackage{xcolor}%
\usepackage{textcomp}%
\usepackage{manyfoot}%
\usepackage{booktabs}%
\usepackage{algorithm}%
\usepackage{algorithmicx}%
\usepackage{algpseudocode}%
\usepackage{listings}%

\author*[1,2]{\fnm{Arianna} \sur{Trozze}}\email{arianna.trozze@ucl.ac.uk}

\author[3]{\fnm{Bennett} \sur{Kleinberg}}\email{bennett.kleinberg@tilburguniversity.edu}
\equalcont{These authors contributed equally to this work.}

\author[2,4]{\fnm{Toby} \sur{Davies}}\email{T.Davies@leeds.ac.uk}
\equalcont{These authors contributed equally to this work.}

\affil*[1]{\orgdiv{Department of Computer Science}, \orgname{University College London}, \orgaddress{\street{Gower Street}, \city{London}, \postcode{WC1E 6EA}, \country{United Kingdom}}}

\affil[2]{\orgdiv{Department of Security and Crime Science}, \orgname{University College London}, \orgaddress{\street{35 Tavistock Square}, \city{London}, \postcode{WC1H 9EZ}, \country{United Kingdom}}}

\affil[3]{\orgdiv{Department of Methodology \& Statistics}, \orgname{Tilburg University}, \orgaddress{\street{Warandelaan 2}, \city{Tilburg}, \postcode{5037 AB}, \country{Netherlands}}}

\affil[4]{\orgdiv{School of Law}, \orgaddress{\street{The Liberty Building, University of Leeds}, \city{Leeds}, \postcode{LS2 9JT}, \country{United Kingdom}}}

\title{Large Language Models in Cryptocurrency Securities Cases: Can a GPT Model Meaningfully Assist Lawyers?}

\abstract{Large Language Models (LLMs) could be a useful tool for lawyers. However, empirical research on their effectiveness in conducting legal tasks is scant. We study securities cases involving cryptocurrencies as one of numerous contexts where AI could support the legal process, studying GPT-3.5’s legal reasoning and ChatGPT's legal drafting capabilities. We examine whether a) GPT-3.5 can accurately determine which laws are potentially being violated from a fact pattern, and b) whether there is a difference in juror decision-making based on complaints written by a lawyer compared to ChatGPT. We feed fact patterns from real-life cases to GPT-3.5 and evaluate its ability to determine correct potential violations from the scenario and exclude spurious violations. Second, we had mock jurors assess complaints written by ChatGPT and lawyers. GPT-3.5’s legal reasoning skills proved weak, though we expect improvement in future models, particularly given the violations it suggested tended to be correct (it merely missed additional, correct violations). ChatGPT performed better at legal drafting, and jurors’ decisions were not statistically significantly associated with the author of the document upon which they based their decisions. Because GPT-3.5 cannot satisfactorily conduct legal reasoning tasks, it would be unlikely to be able to help lawyers in a meaningful way at this stage. However, ChatGPT's drafting skills (though, perhaps, still inferior to lawyers) could assist lawyers in providing legal services. Our research is the first to systematically study an LLM's legal drafting and reasoning capabilities in litigation, as well as in securities law and cryptocurrency-related misconduct.}

\begin{document}

\maketitle

\noindent \textbf{Keywords:} Cryptocurrency, securities law, artificial intelligence (AI), large language models (LLMs), ChatGPT

\section{Introduction}

One of the most promising advances in the field of AI in recent months has been improvements in Large Language Models (LLMs), and one potential application of these is in conducting legal tasks. The public release of a user-friendly LLM in the form of ChatGPT in November 2022 has amplified discussions surrounding such applications. Already, companies have sought to use AI-based tools to draft and file class action or other lawsuits  \citep{kahana_chatgpt_2023}. Similarly, attorneys have experimented with using AI-based tools to automate some of their work (with mixed success, with one attorney submitting a ChatGPT-generated brief citing fictitious cases \citep{martinson_chatgpt_2023}). While the potential for using existing AI tools in the legal field has been documented (see, for example, \citep{perlman_implications_2022, nay_large_2023}), so far, empirical research on its usefulness in practice is sparse. This study fills this gap by systematically testing GPT-3.5's legal reasoning and ChatGPT's drafting capabilities.

The arguments about the potential of legal AI apply to various areas of law. However, in this study, we focus on securities cases involving cryptocurrencies as an example of the many possible contexts where AI might assist in the legal process. We focus on this area as it is one where enforcement attorneys face significant resource constraints that have been chronicled in both the academic and governmental literature \citep{kedia_secs_2011, bonsall_wearing_2019, holzman_when_2023, us_securities_and_exchange_commission_enforcement_2017, us_department_of_justice_report_2022}. This suggests cases in this area could be well-served by AI-based assistance. In the absence of government enforcement as a result of these resource limitations, some victims of cryptocurrency securities violations have also sought justice through securities class action cases;\footnote{\footnotesize{A class action case is a lawsuit brought by someone on behalf of a larger group of plaintiffs who have suffered the same damage as the ``lead plaintiff'' who represents them and their interests.}} as of March 1, 2023, there had been 77 such cases filed \citep{stanford_law_school_securities_2023}. 

Against this background, we seek to answer the following research questions:

\begin{enumerate}
    \item Given a fact pattern from a securities case involving cryptocurrencies, can GPT-3.5 accurately discern which laws are potentially being violated?
    \item Is there a difference in juror decision-making between complaints written by a lawyer and those written by ChatGPT in the context of cryptocurrency securities class action lawsuits?
\end{enumerate}

We conduct our research using the popular, publicly available LLM from OpenAI, GPT-3.5, as an operalization of an LLM, evaluating its ability to identify laws potentially being violated from fact patterns in real-life cases (while not alleging erroneous violations). For our second research question, we had mock jurors assess pleadings written by ChatGPT (the May 27, 2023 version, based on GPT-3.5 and further-fined tuned using human feedback) and lawyers. The current study is the first to evaluate an LLM's legal drafting capabilities in a systematic way and to explore its legal reasoning skills in litigation (and, more specifically, the first to test it in the areas of securities law or wrongdoing involving cryptocurrencies). We evaluate these LLMs' capabilities in two basic parts of the legal process---legal drafting (specifically the first document filed in a case) and legal reasoning in a litigation context---to gain insight into whether LLMs have any potential to meaningfully assist human lawyers. Lawyers spend up to 60\% of their time on legal drafting and legal reasoning and drafting are inextricably linked \citep{litera_litera_2024, gale_legal_1979}. This suggests that LLMs assisting with these tasks would be particularly useful to lawyers. Furthermore, if the LLMs cannot handle these fundamental tasks, they are unlikely to be able to conduct any of the more advanced tasks required of lawyers. 



\subsection{Large language models and their use in legal contexts}

\subsubsection{Introduction to LLMs}

LLMs are advanced AI systems that process human language (referred to as ``natural language'' in this context). This means they can generate and respond to human text, with recent systems appearing to be capable of doing this in a manner that is near-imperceptibly distinct from humans. The state of the art employs a deep learning technique called the Transformer architecture to accomplish this \citep{topal_exploring_2021}. LLMs have advanced rapidly in recent months, most prominently with OpenAI's release of GPT-3 in November 2022 \citep{choi_chatgpt_2023}. GPT stands for ``Generative Pre-trained Transformer'', and was first introduced by \cite{NEURIPS2020_1457c0d6}. What differentiates LLMs from other natural language processing (NLP) models is the massive amount of data on which they are pre-trained to conduct these tasks and the complexity of this Transformer architecture; GPT-3, for example, has 175 billion parameters \citep{choi_chatgpt_2023}. LLMs' colossal amount of training data (though it should be noted that OpenAI's training data is proprietary) means they perform well at natural language tasks in a variety of contexts, whether or not they have been trained on a particular task specifically. For a more detailed introduction to LLMs see \citep{zhao2023survey} and \citep{hadi_large_2023}.

\subsubsection{OpenAI's LLMs}

Alongside GPT-3, OpenAI also released a user-friendly public interface based on the model, called ChatGPT, which was further fine-tuned using Reinforcement Learning with Human Feedback \citep{choi_chatgpt_2023}. This gave users the opportunity to experiment with an LLM in various contexts, without the technical expertise previously required. Since November 2022, OpenAI has released several variations of the GPT-3 model, as well as of the subsequent iterations, GPT-3.5 and GPT-4. In addition to interacting with the latest fine-tuned version of GPT-\textit{n} through the user interface, more tech-savvy users can access specific models using OpenAI's API. These vary by the period over which training data was collected, the amount of input and output they can handle, and some are optimized for more specific tasks like writing computer code. Full details of the current, most recent releases can be found here: \url{https://platform.openai.com/docs/models}. In this study, we use GPT-3.5 models available publicly at the time of data collection, namely, {\tt{gpt-3.5-turbo-0301}} (the basis of ChatGPT at the time) and {\tt{text-davinci-003}}.

\subsubsection{Using OpenAI's LLMs}

To use any of OpenAI's LLMs, the user must provide what is called a ``prompt'' from which the model will generate natural language. An example of such a prompt could be ``Please identify five plants native to New Mexico.'' The nature of LLMs and their proprietary training data means it is far from clear which prompts will be most effective in accomplishing a given task or producing a desired output; the range of possible prompts is as broad as language itself, and there is no established formula for composing them. The process of developing a suitable prompt for an LLM is referred to as ``prompt engineering''. 

Developers have the option of fine-tuning the model by training it on additional data relevant for a specific, desired task. Not doing this is referred to as ``zero-shot'' learning, i.e., using the model as pre-trained \citep{allingham_simple_2023}. ``Few-shot'' learning refers to providing the model with some examples in the prompt \citep{openai_fine-tuning_nodate}. In the example provided above, a few-shot prompt could be:\footnote{\footnotesize{We asked ChatGPT to generate a few-shot prompt for this task, which we adapted and present here.}}

\begin{quote}
    Task instructions: 
    \\
    \\
    1. Flora native to California (provided): [California native plant 1], [California native plant 2], [California native plant 3]
    \\
    \\
    2. Flora native to Texas (provided): [Texas native plant 1], [Texas native plant 2]
    \\
    \\
    New task: New Mexico native flora
    \\
    \\
    Please use your knowledge of native plants of California and Texas to provide five distinct native plant species in New Mexico. 
    \\
    \\
    Desired New Mexico plant species (to be filled in by model): [Plant name], [Plant name], [Plant name], [Plant name], [Plant name]
\end{quote}

When using OpenAI's models through their API, users have several additional parameters they can specify. All of these settings can be found at \url{https://platform.openai.com/docs/api-reference}, however, we highlight two. The first is the {\tt{temperature}}, which determines how random (as opposed to deterministic) the output of the model is. Possible temperature values range from 0 to 2, with higher values adding more randomness to the outputs. Users must also set the maximum number of ``tokens''---a term used in natural language processing (NLP) to refer to elements of text (e.g., words or characters)---they will allow the model to use in its output. OpenAI charges users of its API based on the amount of tokens used in the input and output of a prompt, with different models allowing different volumes of tokens to be used \citep{openai_api_nodate}.

\subsection{LLMs' use in legal tasks}

Given GPT-3's fairly recent release, there is limited empirical research on its use in the legal context, and that which does exist is generally not yet peer-reviewed. Prior research has explored comparatively simpler NLP models' usefulness in law, exploring tasks like judgment extraction or indexing large bodies of case law \citep{yu_legal_2022}. Given the magnitude of recent advancements in this field, however, we review here only research involving GPT-3 or higher.

\subsubsection{Legal tasks}

Existing research in the field of legal applications of LLMs does not necessarily focus on litigation (let alone government enforcement), but rather LLMs' general knowledge of the law and their ability to apply it. Where tasks related to litigation are included (for example, in \citep{iu_chatgpt_2023}), the focus has been more general, rather than applied to a particular aspect of U.S. civil or criminal procedure. Most commonly, research in this field has involved testing LLMs' performance on legal exams, for example, in taking a bar or law school exam. \cite{bommarito_ii_gpt_2022} had GPT-3.5 take the multi-state multiple choice U.S. bar exam, on which it earned a 40.3\% score, compared with a 25\% baseline score for guessing. \cite{yu_legal_2022} asked GPT-3 to conduct legal reasoning tasks that are part of the Japanese Bar exam. They wrote prompts corresponding to existing legal reasoning techniques for a task that requires the lawyer to determine whether a hypothesis is true or false based on a given premise. Finally, \cite{choi_chatgpt_2023} asked ChatGPT to complete various law school exams involving both essay and multiple choice questions; it earned a C+ score. 

Similarly to our research, \cite{nay_large_2023} and \cite{blair-stanek_can_2023} gave versions of the {\tt{text-davinci-003}} model legal reasoning tasks outside of an examination setting, asking it to apply statutes to answer questions and evaluating its performance systematically. \cite{nay_large_2023} used multiple choice questions for this task. \cite{blair-stanek_can_2023} used the StAtutory Reasoning Assessment (SARA) data set for their task, finding that GPT-3 performed significantly better than BERT\footnote{\footnotesize{BERT is a less advanced LLM. For a full introduction to BERT, see \citep{koroteev_bert_2021}.}}, but ``performed at chance (0.5) or worse in the zero-shot tests where there was no statute included.'' They noted that, in particular, GPT-3 displayed incorrect knowledge of U.S. tax code. When given synthetic statutes, GPT-3 performed even worse, which they claim raises ``doubts about GPT-3’s ability to handle basic legal work'' \citep{blair-stanek_can_2023}. 

\cite{savelka_unlocking_2023} employed GPT-3.5 in a task involving the semantic annotation of legal documents. They found that it performed well---and, in fact, better than fine-tuned versions of BERT and a Random Forest Model---with scores of $F1 = 0.73$ in identifying key topics in court opinions, $F1 = 0.86$ for the same in contracts, and $F1= 0.54$ in statutes and regulations. 

\cite{iu_chatgpt_2023} and \cite{perlman_implications_2022} conducted preliminary exploration of ChatGPT’s general drafting and legal analytical capabilities. For example, \cite{iu_chatgpt_2023} asked ChatGPT to prepare cross-examination questions for a case and draft a skeleton argument based on analysis of a complex fact pattern, among other tasks. They did not prompt the model or evaluate their results in a systematic way. They did, however, suggest ChatGPT performed well at this task, though not as well as a lawyer. These authors also used ChatGPT to assist them in drafting their papers.

\subsubsection{Models}

In terms of the particular LLMs tested, three studies \citep{bommarito_ii_gpt_2022, savelka_unlocking_2023, blair-stanek_can_2023} used {\tt{text-davinci-003}}, as we do in answering our first research question. \cite{nay_large_2023} used various models, noting that GPT-4 (the state of the art at the time) performed best and that they actually used GPT-4 to help grade their answers. \cite{yu_legal_2022} used GPT-3. 

\subsubsection{Areas of law}

The most common area of law on which prior research focused is tax law. \cite{nay_large_2023}, \cite{blair-stanek_can_2023}, and \cite{choi_chatgpt_2023} all included tax law in their studies. \cite{bommarito_ii_gpt_2022} note that GPT-3.5 passed the evidence and torts section of the bar exam, and \cite{choi_chatgpt_2023} also included an examination paper on torts in their work. In addition to these two subjects, \cite{choi_chatgpt_2023} also examined ChatGPT on constitutional law, federalism and separation of powers, and employee benefits. \cite{savelka_unlocking_2023} researched GPT-3.5's capabilities in relation to veterans' affairs cases, public health cases, and general commercial legal contracts. 

\subsubsection{Prompt engineering and parameter selection}

Much previous research (though we note most had not been released at the time of our study design) focused on prompt engineering in legal contexts. \cite{yu_legal_2022} found that using the legal reasoning technique ``Issue, Rule, Application, and Conclusion'' (IRAC) in their prompts performed best. Some research \citep{bommarito_ii_gpt_2022, savelka_unlocking_2023} found that zero-shot prompting proved more effective compared to fine-tuned models, and \cite{choi_chatgpt_2023} determined that, generally, the simplest prompts performed best. 

However, this is contradictory to other findings, such as those of \cite{nay_large_2023} and \cite{blair-stanek_can_2023}, which used zero-shot performance as a baseline. \cite{nay_large_2023} also explored chain-of-thought (CoT) prompting and few-shot prompting, and combined these approaches. CoT prompting involves asking an LLM to ``think step-by-step'', which \cite{kojima_large_2023} previously found to improve performance (though not specifically in legal tasks). \cite{nay_large_2023} found that CoT only improved the results for GPT-4 and that providing relevant legal texts in the prompts moderately improved the results. Few-shot prompting led to better results for GPT-4 but not for other models. 

\cite{blair-stanek_can_2023} used up to 10-shot CoT prompting and also found, again contrary to \cite{kojima_large_2023}'s results, this was unhelpful in their experimental context. In general, their prompts were much more complex: for example, in their ``4-shot dynamic prompting'' context, they included the four most similar training cases and labelled the text of the case (which they also provided in full) with the ``premise'', the ``hypothesis'', and the ``answer''. Having to prepare prompts in this way may not actually save lawyers time. Notably, \cite{choi_chatgpt_2023} specifically asked ChatGPT not to ``fabricate'' any facts, and it followed this instruction. Overall, these findings suggest prompt engineering is much more of an art than an exact science. 

\citeauthor{perlman_implications_2022}'s (\citeyear{perlman_implications_2022}) was the only study to ask ChatGPT to draft a complaint, which, unsurprisingly, resulted in a very short complaint being drafted. We note that neither they, nor \cite{iu_chatgpt_2023}, engaged in any documented attempts to refine their prompts. \cite{perlman_implications_2022} used the following prompt: 

\begin{quote}
    ``Draft a legal complaint for a Massachusetts state court by John Doe against Jane Smith for injuries arising out of a car accident on January 1, 2022 caused by Jane Smith at the intersection of Tremont Street and Park Street in Boston. The complaint should specify that Jane Smith failed to stop at a red light and caused John Smith serious injuries.''
\end{quote}

For drafting essays, \cite{choi_chatgpt_2023} asked ChatGPT to write essays section by section; we use a similar prompting approach in this study. 

\subsubsection{Consideration of LLMs' training data}

Since LLMs have been trained on large corpora of real-world text data, it is possible that some legal sources are present in the training data and therefore have previously been ``seen'' by the models, which the literature considers at length.\footnote{Since the training data for most LLMs, including ChatGPT, is proprietary, this cannot be confirmed directly.} As noted in some studies \citep{bommarito_ii_gpt_2022, nay_large_2023, blair-stanek_can_2023}, this represents a potential threat to the validity of research of this nature since a model may successfully complete a task simply by reproducing known material, rather than by ``interpreting'' its meaning. \cite{bommarito_ii_gpt_2022} identified behavior which appeared to indicate that there were some legal sources in its training. In their research, \cite{nay_large_2023} and \cite{blair-stanek_can_2023} specifically used synthetic cases to ensure the LLM had not been been trained on them. Notably, \cite{blair-stanek_can_2023} found that GPT-3 was unlikely to have been trained on the complete U.S. tax code because it showed incorrect knowledge thereof. They note that, ``given a U.S. Code citation, GPT-3 can recite plausible but incorrect statutory language'' \citep{blair-stanek_can_2023}. 

\subsection{Cryptocurrency securities violations and related law}

As discussed above, the use of LLMs might be particularly useful to lawyers in the cryptocurrency securities context due to documented resource limitations faced by enforcement attorneys in this area of law. We use examples from this domain to perform our study. In this section, we provide a brief outline of the legal context for cryptocurrency securities violations.

\subsubsection{U.S. securities laws and securities class action lawsuits}

A security is ``an investment of money in a commercial, financial, or other business enterprise, with the expectation of profit or other gain produced by the efforts of others'' \citep{ninth_circuit_jury_instructions_committee_manual_2023}. Specific types of securities include stocks, bonds, and investment contracts \citep{ninth_circuit_jury_instructions_committee_manual_2023}. In the U.S., at the federal level, the offer, sale, and purchase of securities is governed by the following laws: Securities Act of 1933 (``Securities Act''); Securities Exchange Act of 1934 (``Exchange Act''); Investment Company Act of 1940; Investment Advisers Act of 1940; Sarbanes-Oxley Act of 2002 (``Sarbanes-Oxley''); and Dodd-Frank Wall Street Reform and Consumer Protection Act of 2010 (``Dodd Frank'') \citep{practical_law_securities_litigaiton__white_collar_crime_securities_2023}. The U.S. Securities and Exchange Commission (SEC) and the Financial Industry Regulatory Authority (FINRA) are responsible for civil securities law enforcement in the U.S. In addition to violations of the aforementioned regulations, the SEC has the power to charge ``securities fraud based on negligent conduct'' and ``aiding and abetting securities fraud'' \citep{practical_law_securities_litigaiton__white_collar_crime_securities_2023}. 

Private securities cases usually take the form of class actions. Private plaintiffs enjoy rights of action under the Securities Act, the Exchange Act, and Sarbanes-Oxley. The most common bases for such suits when they involve cryptocurrencies are are Section 10(b) of the Exchange Act and SEC Rule 10b-5 (securities fraud) and Sections 5(a), 5(c), and 12(a)(1) of the Securities Act (unregistered securities offering, and liability of the offerer or seller thereof) \citep{practical_law_securities_litigaiton__white_collar_crime_securities_2023, noauthor_civil_nodate, noauthor_prohibitions_nodate}. For a full overview of U.S. civil procedure and the procedure for class action suits as set out in the PSLRA, see Appendix \ref{civproced}.

\subsubsection{Introduction to cryptocurrencies}

Cryptocurrencies are a store of value initially conceived in 2008 by a developer (or group) using the pseudonym Satoshi Nakamoto \citep{home_office_factsheet_2023, nakamoto_bitcoin:_2008}. Nakamoto proposed the first cryptocurrency, Bitcoin, as a peer-to-peer ``digital representation of value'' which would enable direct transactions between individuals without the need for an intermediary financial institution \citep{home_office_factsheet_2023}. Bitcoin uses a novel combination of existing cryptographic\footnote{\footnotesize{Cryptography refers to the science of secure communications, namely ways to ensure communications remain confidential, authentic, tamper-resistant, and only comprehensible by intended participants. It is used for securing various types of information and permeates several facets of everyday life, from sending encrypted text messages to online banking.}} concepts called a blockchain, which serves as a decentralized ledger of all transactions \citep{narayanan_blockchains_2018, moffett_cftc_2022}. Transactions are secured through cryptography, with participants verifying transactions by solving computationally complex cryptographic problems in a process called ``mining''. As part of this mining process, valid transactions are combined into ``blocks'' which are added to the blockchain. They are ``chained'' to previous blocks in a way that can be cryptographically verified, meaning that nefarious actors cannot alter this digital ledger, thereby ensuring the system’s security \citep{nakamoto_bitcoin:_2008}. For a more complete introduction to Bitcoin and cryptocurrencies, see \citep{narayanan_bitcoin_2016}.

\subsubsection{Cryptocurrencies as securities}

Since Bitcoin was first made available in 2009, the cryptocurrency ecosystem has greatly expanded, with more than 26,000 cryptocurrencies in existence today \citep{coinmarketcap_cryptocurrency_nodate}. These may exist on their own blockchains (or other forms of decentralized ledger technology) or can be created on existing blockchains, such as the Ethereum blockchain. In the U.S., both the Commodity Futures Trading Commission (CFTC) and the SEC have claimed to have regulatory control over cryptocurrencies in certain situations. In many cases, certain cryptocurrencies' primary purpose has been determined to be as an investment contract, thereby classifying them as securities \citep{moffett_cftc_2022}. As discussed above, the SEC is responsible for securities regulation and enforcement. Though a full discussion of the SEC's reasons for considering certain cryptocurrencies as securities is outside of the scope of this paper, we give a brief overview. For a full consideration thereof, see \citep{moffett_cftc_2022}.

The SEC and the U.S. federal courts have traditionally based their analysis as to whether a cryptocurrency constitutes a security on \textit{SEC v. W.J. Howey Co.} (now colloquially referred to as the ``Howey test''). In this Supreme Court case, an investment contract needed to meet the following criteria: it had to be ``(1) an investment of money, (2) in a common enterprise, (3) with the expectation of profits, (4) to be derived from the entrepreneurial or managerial efforts of others'' \citep{moffett_cftc_2022}.

Based on the Howey test, Bitcoin is not a security (specifically because it does not satisfy requirements two and four discussed above), but one could argue that many other cryptocurrencies (such as those featured in the cases we explore below) are. Notably, it is possible that a cryptocurrency’s status as a security can change over time \citep{moffett_cftc_2022}. 

\section{Methods}

This study seeks to discover whether (1) when presented with a fact pattern from a securities enforcement case involving cryptocurrencies, GPT-3.5 can determine which laws are potentially being violated, and (2) whether, for securities class action lawsuits involving cryptocurrencies, there is a difference in juror decision-making based on complaints written by a lawyer compared with those written by ChatGPT. 

We used versions of OpenAI's GPT-3.5 models to answer both questions. For the first, we prompted GPT-3.5 to discern which laws were potentially being violated, when presented with fact patterns from real-life cases. We then calculated various evaluation metrics to test its performance, using the allegations in the cases from which we extracted the fact patterns as our ``ground truth'' for correctly identified violations. The closer GPT-3.5 got to identifying all of the counts in the actual complaint (and without alleging erroneous charges), the better we considered its performance at this task. 

To answer our second research question, we presented mock jurors with pleading documents written by either ChatGPT or a human lawyer and asked them to answer various questions. The performance of ChatGPT in legal drafting was considered to be adequate if jurors reached the same decisions when presented with the AI-drafted complaint as they did for that written by a human lawyer. Ordinarily, jurors would not evaluate complaints, but rather would make a decision based on lawyers' arguments and evidence presented (mostly verbally) at trial. However, in this study, we asked jurors to evaluate the complaint directly, assuming the facts presented therein are true. In legal procedure, complaints would typically be read and evaluated by judges and lawyers, and so, it would be useful for lawyers to review such complaints that were generated; however, in this study, we did not have access to a sufficient number of lawyers to produce a robust evaluation of ChatGPT's legal drafting skills. Ultimately, we opted to provide detailed jury instructions to a larger number of laypeople to make decisions based on a legal document. 

In our study, we used zero-shot transfer.\footnote{\footnotesize{Zero-shot transfer means that the LLM has neither been given relevant examples in the prompt, nor further fine-tuned (i.e., trained on many examples to produce a model which will accept either zero-shot or multi-shot prompts). However, because large-scale pre-trained models could happen to incorporate some relevant examples in their training data, this is distinct from zero-shot learning \citep{allingham_simple_2023}.}}  In developing prompts to explore each research question, we tested them on a general securities enforcement case, \textit{SEC v. Kaplan, et al.}, case number 2:23-cv-01648, in the U.S. District Court for the Eastern District of New York. 

\subsection{GPT-3.5's ability to discern violations of U.S. law}

We examined GPT-3.5’s ability to determine, when given a fact pattern corresponding to a case, and prompted with a particular legal reasoning technique, which U.S. laws were potentially being violated. Civil complaints contain a section detailing the facts of the case and describing the defendants' alleged conduct. For each case we study, we provided the information (verbatim) in the factual allegations section to GPT-3.5. For a more detailed introduction to legal complaints, see Appendix \ref{civproced}.

\subsubsection{Choice of model}

To answer this research question, we used the GPT-3.5 {\tt{text-davinci-003}} model, which is trained on data collected prior to June 2021. Aside from newer models being more expensive to run, their training data sets are more recent. For this research question, we wanted to ensure that the model had no prior knowledge of our cases (i.e., that these cases could not be present in its training data). We used OpenAI’s native Python wrapper to execute our API calls. 

\subsubsection{Case selection}

In selecting our cases, we first identified all cryptocurrency securities class action and SEC enforcement cases which were filed after June 2021 (and before our data collection date, on March 24, 2023). We included all SEC cases under the heading ``Crypto Assets'' on the SEC's Crypto Assets and Cyber Enforcement Actions page \citep{us_securities_and_exchange_commission_crypto_2023}. This list includes all SEC enforcement actions that involve cryptocurrencies. For the class action cases, we collected all cases included in the Stanford Securities Class Action Clearinghouse Filings Database listed under ``Cryptocurrency Litigation'' in their ``Current Trends'' section \citep{stanford_law_school_securities_2023}. Again, these are securities class action cases that involve cryptocurrencies.  This gave us an initial set of 34 class action cases and 43 enforcement actions. Next, we extracted the factual allegations sections of the cases' respective complaints. Because the maximum number of tokens for the {\tt{text-davinci-003}} model is 4,097, we excluded all cases whose facts sections exceeded this number of tokens (using OpenAI’s Tokenizer tool to calculate the number of tokens) \citep{openai_tokenizer_nodate}. This resulted in the exclusion of 50 cases. We also excluded any administrative SEC cases ($n=9$). 

There were two pairs of cases which were filed separately, but with nearly identical facts sections, due to lawyers having chosen not to join the defendants. For each of the two pairs, we chose to include whichever case had a more detailed facts section in our final data set and excluded the other. 

This left us with a final set of 20 cases, eight of which are class action cases. More details of our case selection process, including reasons for excluding specific cases, can be found in Appendix \ref{cases1}. 

\subsubsection{Prompt design}

To craft our prompt for this aspect of our study, we used ChatGPT's online user interface. Our full prompt design process—tested on our sample case, \textit{SEC v. Kaplan, et al.}—is detailed in Appendix \ref{prompt1}, though we highlight selected considerations here. Following \citeauthor{yu_legal_2022}'s (\citeyear{yu_legal_2022}) research, we requested ChatGPT use a specific legal reasoning technique (IRAC). Second, as we conducted our prompt engineering, it was clear we needed to be as specific as possible in our prompt. Our final prompt, therefore, defined the jurisdiction in which the facts occurred or were connected (based on where the real complaint was filed; this also communicates to the LLM that we are interested in U.S. law for this task), indicated that the violations should be federal and civil (as opposed to state or criminal laws), and that the LLM should state the specific section of any statute potentially violated.

We arrived at the following, final prompt:

\begin{quote}
``The following text is from the \textbackslash``factual allegations\textbackslash'' section of a complaint filed in the [\textit{jurisdiction; in the sample case, Eastern District of New York}]. Based on the facts in this text, please identify which federal civil law(s) and section thereof the defendant(s) violated. Please use the following method of legal reasoning to come up with the allegations: Issue, Rule (including the specific statute and section thereof), Application, Conclusion: [\textit{text from factual allegations section}]''
\end{quote}

\subsubsection{Input pre-processing and cleaning}

Before running our prompt on our chosen cases, we pre-processed and cleaned our input text. We manually removed the following from the complaint text: paragraph numbers; footnotes; references; pictures; indications that emphasis was added; background information (for example, an introduction to cryptocurrency); and extra spaces. We also added the appropriate escape character (``\textbackslash'') before quotation marks and apostrophes. 

\subsubsection{Parameters and execution}

Before applying our prompt to our final set of cases, we considered the appropriate temperature setting for our model. Ultimately, we wanted the model to follow our instructions closely, so we chose 0.2 as our temperature.\footnote{\footnotesize{We chose 0.2 instead of 0 because, while we wanted a highly deterministic output, we wanted some randomness to allow for some variability in our output. During our prompt engineering process, when we set the temperature to 0, each iteration yielded the exact same output.}} We also experimented with a temperature of 1 (for a more random output), however, we found that this resulted in little change in the output. We used the same temperature of 0.2 as we sought to identify the baseline performance of ChatGPT for a task of this kind (rather than exploring how we can fine-tune its parameters for better performance). For each case, we ran the model five times (each with the same temperature of 0.2). 

The {\tt{max\_tokens}} parameter specifies the maximum length of text to be generated: we set this to the total number of tokens this model accepts (4,097) minus the number of tokens provided in the prompt for each case, thereby allowing the maximum possible size for each output. We also recorded the number of tokens used for completion in each output for each of our cases; this can be found in Appendix \ref{results1}.

\subsubsection{Evaluation}

To evaluate GPT-3.5's performance, we measured the extent to which the violations it identified matched those in real cases. For this purpose, we used the allegations in the real-life complaint as our ``ground truth''. For each case, multiple violations apply in the real case, and GPT-3.5 can also identify multiple violations, but the number of violations may not be equal. Therefore, we employed performance metrics to measure the extent to which GPT-3.5 identified the correct violations or identified incorrect ones. We note that there is some limitation to automatically considering any additional charges suggested by GPT-3.5 as ``incorrect'', as lawyers sometimes make mistakes or decide not to include certain charges for other, strategic reasons. However, for the purposes of this study, we consider the lawyers' charges as ``correct'' based on the facts in question.

With that in mind, we adapted the definitions of true positives, false positives, and false negatives for our context to calculate performance metrics. In this case, the number of true positives (TP) was the number of claims GPT-3.5 correctly identified. The number of false positives (FP) was the number of incorrect, additional violations suggested. The number of violations GPT-3.5 missed constituted the false negatives (FN). The number of true negatives was not applicable in our analysis because it would not be meaningful to count every possible law that does not apply in this context.

\begin{figure}
    \centering
    \includegraphics[width=12cm]{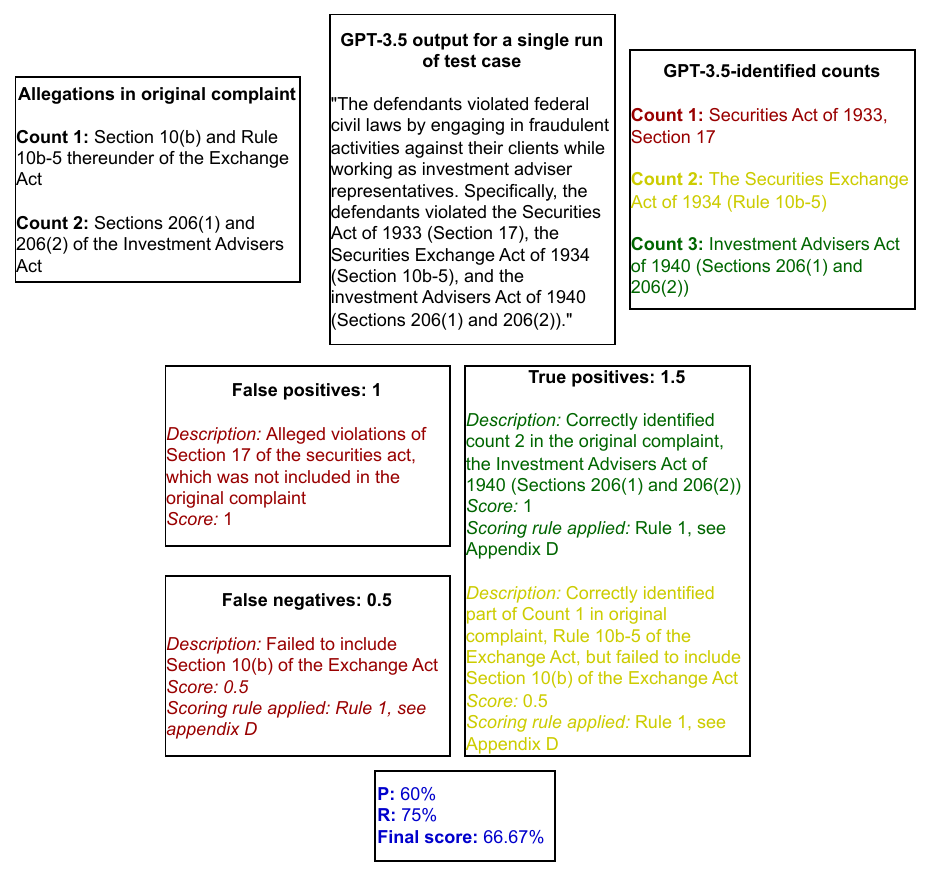}
    \caption{Scoring GPT-3.5 output for one run of test case}
    \label{fig2}
\end{figure}

We use adapted definitions because we allow for partial counts of these scores to capture the nuance of the legal context. To be considered ``correct'', GPT-3.5 needed to identify all elements of each legal claim. Otherwise, it would receive a partial score, as per the following rules: 

\begin{itemize}
    \item Rule 1: In cases where violations are almost always alleged together (for example, Section 10(b) of the Exchange Act and Rule 10b-5 thereunder), we counted this as a single violation and scored accordingly. However, GPT-3.5 only identified, for example, a Rule 10b-5 violation, but not a violation of Section 10(b), we would score this as 0.5.\footnote{\footnotesize{There is some further nuance to this in our sample case (see Figure \ref{fig2}) because we do not specify in our prompt whether the case in question would be a government enforcement or private matter and there is no private right to action under Section 10(b) unless the company is publicly listed. However, while some courts have ruled there is a private right of action under Section 17(a), the generally accepted view is that this is not the case (and, besides, GPT-3.5 did not specify any part of Section 17 in its output for our sample case) \citep{legal_information_institute_1933}. This implies that GPT-3.5 either a) assumed this would be a government enforcement case and, therefore, charging only 10b-5 without 10(b) is unusual, or b) it assumed this was a private case and the Section 17 finding was wrong. Our scoring system, therefore, better captures these nuances. While it is, of course, not infallible and still assumes the charges in the real complaint are the ground truth, it would be impractical to have several lawyers analyze each output from GPT-3.5 to assign it a score.}}
    \item Rule 2: We also awarded 0.5 points if the output included the correct law, but failed to include the specific section the defendants violated. So, for example, if the output suggested violations of the Securities Act, but did not specify that the allegations were of Sections 5(a) and 5(c) thereof, 0.5 points would be awarded. However, if the output merely stated that violations of ``federal securities laws'' occurred, it was given a score of 0.
    \item Rule 3: In cases where the complaint charged Sections 5(a) and 5(c) of the Securities Act and the output only included Section 5 thereof (overall), we considered this as a true positive (because Section 5 overall would include Sections 5(a) and 5(c)).
    \item Rule 4: For the purpose of calculating true positives and false negatives, where the complaint charged different counts of the exact same allegations in the complaint (usually just for different defendants), we counted them as a single violation. That being said, there was one case (\textit{Securities and Exchange Commission v. Arbitrade Ltd., et al.}, 1:22-cv-23171, S.D. Fla.), where one claim was for Rule 10b-5 under the Exchange Act and another for 10b-5(c) (against a different defendant). However, since one of the charges did include all of Rule 10b-5, we counted these both as a single violation.
    \item Rule 5: We did not infer any charges if the output failed to reference the appropriate statute (i.e., if it included only ``unregistered securities'', we did not assume it meant a violation of Section 5 or Section 12(a)(1) of the Securities Act).
    \item Rule 6: Finally, the prompt specifically requested violations of federal laws. Some outputs included state law violations. Because this was contrary to the prompt’s instructions, these were automatically considered as false positives for the sake of scoring. 
\end{itemize}

Figure \ref{fig2} depicts our scoring process for the output for one run of our test case.

After applying these rules manually to the output, we evaluate GPT-3.5's performance at this legal reasoning task. We used our previously defined, adapted concepts of true positives, false positives, and false negatives to calculate precision and recall (similarly to how \cite{ting_precision_2010} applied such metrics to a document retrieval system, as we were, essentially, asking GPT-3.5 to retrieve relevant laws from a finite set of laws). Precision ($P$) is the total number of correct charges identified (true positives), divided by the total number of charges GPT-3.5 identified (true positives plus false positives):

\[P = \frac{\text{TP}}{\text{TP + FP}}\]

Recall ($R$) is total number of correct charges identified (true positives) divided by the total number of possible correct charges (true positives plus the false negatives).

\[ R = \frac{\text{TP}}{\text{TP + FN}}\]

We then use the output of these measures to calculate an $F\text{-}1$ score that balances the precision and recall based on our adapted versions of TP, FP, and FN, as our final score, calculated as follows:

\[ {F\text{-}1} = \frac{\text{$2 \cdot P \cdot R$}}{\text{P + R}}\]

We calculate the average scores for each metric across the five outputs for each of our 20 cases, as well as for all of our outputs ($n=100$). 

The full outputs of the prompts for each of our cases, as well as our calculated scores, are in Appendix \ref{results1}.

\subsection{Differences in juror decision-making based on human vs. AI-written legal complaints}

To answer our second research question, we sought to abstract the legal process to discern whether jurors made different decisions when presented with legal allegations and factual scenarios written by ChatGPT versus those written by human lawyers. We explored this using real-world securities class action cases. 

\subsubsection{Case selection}

In order to more fulsomely evaluate ChatGPT's legal drafting skills, we included cases of varying quality. We wanted to evaluate ChatGPT's drafting ability not only for high quality cases (i.e., those a judge decided had merit to continue through the legal process), but also those of lower quality (i.e., that a judge dismissed). We used class action cases because, overwhelmingly, cases brought by the SEC are successful: in 2022, for example, it won 12 of its 15 trials \citep{us_securities_and_exchange_commission_sec_2022}. The sample of such cases that are ``without merit'' is, therefore, extremely small, potentially limiting the insight that could be gained. 

To select our cases, we first identified all cryptocurrency class action cases, again, using the Stanford Securities Class Action Clearinghouse Filings Database, specifically collecting cases listed under ``Cryptocurrency Litigation'' in their ``Current Trends'' section \citep{stanford_law_school_securities_2023}. Initially, we examined cases which had been filed between ChatGPT’s June 2021 training cut-off date and the date of our research (March 24, 2023). This gave us 34 cases. We sought cases which met the following criteria:

\begin{enumerate}
    \item For ``valid'' cases, the class action cases had to have had a motion to dismiss filed, considered, and, ultimately, denied by the judge.
    \item For our ``meritless'' cases, the class action cases had to have a motion to dismiss filed, considered, and, ultimately, ordered affirmatively by the judge. 
\end{enumerate}

From our initial set of cases, only one case from each category met these criteria, which would not allow us to evaluate ChatGPT’s pleading drafting abilities in a particularly robust way. Therefore, we expanded the temporal range of our search to include all cryptocurrency securities class action cases, the first of which was filed in 2016 \citep{stanford_law_school_securities_2023}. This increased the total number of cases to evaluate to 77. We analyzed these cases (based on their progress as of May 22, 2023), excluding ones which did not meet the criteria. We did not include cases which involved voluntary dismissal or where the motion to dismiss was granted in part. This gave us a set of 30 cases. 

In order to draft our AI-generated complaints, we provided ChatGPT with the facts as presented in Law360\footnote{\footnotesize{\url{https://www.law360.com/}}} articles about the filing of a particular complaint. Law360, a subsidiary of LexisNexis, is a reputable source of legal news that writes stories on major cases summarizing the facts from the complaint when it was filed (see Section \ref{chatgptcomplaints} below). Prior legal research (see \citep{ahmad_outcomes_2022} and \citep{kacperczyk_legal_2024}) has utilized Law360 databases. While these articles are unlikely to be completely representative of the facts that would have been available to plaintiffs and their lawyers when the complaints were originally drafted, these types of legal news articles are the best approximation we have thereof. We excluded any cases for which there was either (a) no Law360 article, or (b) the only articles were about another filing (rather than the complaint). After doing this, nine cases remained in our data set. We note that, in cases where an amended complaint was filed, we used whichever complaint was referenced in the Law360 article as our lawyer-drafted complaint. 

For full details on the selection process, including the reasons for exclusion for each of the cases in the set, see Appendix \ref{cases2}. Our final nine cases can be found in Table
\ref{tab1} below.

\begin{table}[h]
\begin{tabular}{p{5cm} | p{5cm}}
\textbf{Dismissed} & \textbf{Continued} \\ \hline
\textit{Lee, et al. v. Binance, et al.}, case number 1:20-cv-02803, in the U.S. District Court for the Southern District of New York & \textit{Hong, et al. v. Block.One, et al.}, case number 1:20-cv-03829, in the U.S. District Court for the Southern District of New York \\ \hline
\textit{Underwood, et al. v. Coinbase Global Inc.}, case number 1:21-cv-08353, in the U.S. District Court for the Southern District of New York & \textit{Balestra v. Cloud With Me  Ltd.}, case number 2:18-cv-00804, in the U.S. District Court for the Western District of Pennsylvania \\ \hline
\textit{Brola v. Nano, et al.}, case number 1:18-cv-02049, in U.S. District Court for the Eastern District of New York & \textit{Audet, et al. v. Garza, et al.}, case number 3:16-cv-00940, in the U.S. District Court of Connecticut \\ \hline
\textit{Ha v. Overstock.com, et al.},  case number 2:19-cv-00709, in the U.S. District Court for the District of Utah & \textit{Klingberg v. MGT Capital Investments Inc., et al.}, case number 2:18-cv-14380, in the U.S. District Court for the District of New Jersey \\ \hline
& \textit{Davy v. Paragon Coin Inc., et al.}, case number 3:18-cv-00671, in U.S. District Court for the Northern District of California \\ \hline  
\end{tabular}%
\caption{Final set of cases for mock jurors}
\label{tab1}
\end{table}

\subsubsection{ChatGPT-drafted complaints}\label{chatgptcomplaints}

We prompted ChatGPT to draft complaints using a series of zero-shot prompts. For further details on the conventions for drafting a legal complaint, see Appendix \ref{civproced}. It is impossible to give ChatGPT the exact same set of facts that the lawyer had to draft the original complaint because we do not have access to that information. However, we are not testing ChatGPT’s ability to discern which facts, from a set of facts, are most relevant legally, but rather its ability to draft a convincing complaint from a given set of facts.

With that in mind, rather than manually extracting the facts of the case from the complaint, we used the facts as presented in the Law360 summary of the case when it was filed. We gave the Law360 article (verbatim, excluding the case number and counsel information) to ChatGPT and asked it to write each section of a complaint for a particular United States District Court (the one where the real complaint was filed). We found that asking ChatGPT for each section of the complaint individually resulted in more detailed and higher quality outputs (as \cite{choi_chatgpt_2023} also found in asking it to draft essays). When we simply asked ChatGPT to draft the whole complaint (even when the sections and suggested length were specified), it produced a short, sparse complaint. We tested our prompts on our sample case, \textit{SEC v. Kaplan, et al.} The Law360 articles used in our prompts were the following: \citep{perera_identical_2023, guarnaccia_rapper_2019, ruscoe_erratic_2019, sinay_investors_2020, sinay_investors_2020-1, archer_cantor_2016, newsham_70m_2018, kochman_irish_2018, wichert_mcafee-linked_2018, jarvis_nba-highlight_2022, innis_coinbase_2022}. For further details of our prompt design process, see Appendix \ref{prompt2}. 

We note that, for this part of our research, we were unable to use the OpenAI API. When we submitted our prompts to the API, we received an error. We were, ultimately, able to generate the desired text by adding ``for educational purposes only'' to our prompt and using the ChatGPT user interface, specifically the May 24, 2023 GPT-3.5 version, which uses the {\tt{gpt-3.5-turbo-0301}} model. For full details of the original output error and associated prompt design considerations, see Appendix \ref{prompt2}.

Our final prompt format is as follows:

\begin{itemize}
    \item Prompt 1: The following article contains facts summarizing a class action complaint filed in the [\textit{insert venue}]. [\textit{Text from the Law360 article about this complaint being filed}]. For educational purposes only, based on the facts summarized and provided above, please draft the caption of a class action complaint for the [\textit{insert venue}].
    \item Prompt 2: The following article contains facts summarizing a class action complaint filed in the [\textit{insert venue}]. [\textit{Text from the Law360 article about this complaint being filed}]. For educational purposes only, based on the facts summarized and provided above, please draft the preliminary statement of a class action complaint for the [\textit{insert venue}].
    \item Prompt 3: The following article contains facts summarizing a class action complaint filed in the [\textit{insert venue}]. [\textit{Text from the Law360 article about this complaint being filed}]. For educational purposes only, based on the facts summarized and provided above, please draft the jurisdiction section of a class action complaint for the [\textit{insert venue}].
    \item Prompt 4: The following article contains facts summarizing a class action complaint filed in the [\textit{insert venue}]. [\textit{Text from the Law360 article about this complaint being filed}]. For educational purposes only, based on the facts summarized and provided above, please draft the parties section of a class action complaint for the [\textit{insert venue}].
    \item Prompt 5: The following article contains facts summarizing a class action complaint filed in the [\textit{insert venue}]. [\textit{Text from the Law360 article about this complaint being filed}]. For educational purposes only, based on the facts summarized and provided above, please draft the factual allegations section of a class action complaint for the [\textit{insert venue}].
    \item Prompt 6: The following article contains facts summarizing a class action complaint filed in the [\textit{insert venue}]. [\textit{Text from the Law360 article about this complaint being filed}]. For educational purposes only, based on the facts summarized and provided above, please draft the class allegations section of a class action complaint for the [\textit{insert venue}].
    \item Prompt 7: The following article contains facts summarizing a class action complaint filed in the [\textit{insert venue}]. [\textit{Text from the Law360 article about this complaint being filed}]. For educational purposes only, based on the facts summarized and provided above, please draft the legal claims for relief section of a class action complaint for the [\textit{insert venue}].
    \item Prompt 8: The following article contains facts summarizing a class action complaint filed in the [\textit{insert venue}]. [\textit{Text from the Law360 article about this complaint being filed}]. For educational purposes only, based on the facts summarized and provided above, please draft the prayer for relief section of a class action complaint for the [\textit{insert venue}].
    \item Prompt 9: The following article contains facts summarizing a class action complaint filed in the [\textit{insert venue}]. [\textit{Text from the Law360 article about this complaint being filed}]. For educational purposes only, based on the facts summarized and provided above, please draft the jury demand section of a class action complaint for the [\textit{insert venue}].
\end{itemize}

We generated each prompt once. Where ChatGPT failed to generate an appropriate caption section for a specific complaint ($n=2$), we used the caption section generated by another prompt for the same case, because each section generated tended to also include the caption as well. We note that, because we used the ChatGPT user interface, we were unable to control the temperature in this part of the study.

We then compiled each of the sections of the complaint generated by ChatGPT into a single document. We excluded anything generated for other sections besides that specified in the particular prompt (i.e., in the output for the caption prompt, we only transposed the caption of the complaint, rather than any output in the ``preliminary statement'' or other sections, if generated by ChatGPT). 

We removed parts of the complaint where ChatGPT indicated a fact that was not included in the article would need to be added. For example, if the AI-generated text for the ``parties'' section was ``Plaintiff A is an individual who resides in [insert state]'', we kept only ``Plaintiff A is an individual''. We also removed any instances where ChatGPT suggested we insert a particular dollar amount and, if the dates of the class period were not present in the Law360 article, we simply included ``during the class period'' in the text, rather than adding the specific dates. 

Our final ChatGPT-drafted complaints, including that for our sample case, can be found in Appendix \ref{complaints1}.

\subsubsection{Lawyer-drafted complaints}

Naturally, the Law360 articles do not contain all the facts that were included in the original lawyer-drafted complaints, but rather a high-level summary thereof. Therefore, for the purpose of our study, we needed to manually edit the lawyer-drafted complaints and remove all text that referenced facts which were not included in the article, so that the lawyer-drafted versions could be meaningfully compared to the AI-generated ones. 

This was also beneficial in terms of reducing the length of the complaints our mock jurors needed to read to reach their decisions; some of the original complaints were more than 200 pages long, which would be impractical for jurors in this study to read in their entirety. We also removed the following: any charges outside of those on which we asked our mock jurors to make decisions (see section \ref{charges} below); footnotes; indications that emphasis was added; background information on cryptocurrencies; and the case number. 

\cite{willmott_introduction_2018} similarly reduced the length of the case document they used in their mock jury study (in their case, a trial transcript). Though it was a concern, this process did not seem to harm the quality of the lawyer-drafted complaints. For each complaint, we confirmed the document still read well after our deletions and we found we did not need to make any additional amendments. Though we acknowledge that these exclusions pose a risk of reducing the persuasiveness of the lawyer-drafted complaints, given that our mock jurors ultimately considered the charges in the lawyer-drafted complaints to be proven 88\% of the time, this did not seem to be the case.

We note that, in one case, \textit{Audet, et al. v. Garza, et al.}, case number 3:16-cv-00940, in the U.S. District Court of Connecticut, there was an error in the Law360 article, which wrongly referred to ``Hashstakers'' as ``Hashstalkers''. Therefore, we changed all instances of the word ``Hashstakers'' to ``Hashstalkers'' in the lawyer-generated complaint to facilitate comparison with the AI-generated one based on the article. 

We also changed the formatting of these complaints, so all the complaints presented to our mock jurors were formatted consistently. See Appendix \ref{complaints1} for the final versions of our lawyer-drafted complaints, including for our sample case. 

\subsubsection{Mock juror decision-making}

\subsubsection{Selecting overlapping charges} \label{charges}

Based on the contents of the cases, we selected the particular violations on which to ask our participants to make judgements. For the sake of comparability, we needed to look at the same charge for both the ChatGPT-generated version of the complaint and the original, lawyer-drafted one for each case. To avoid introducing further variables into our analysis, we sought to choose the smallest number of charges possible which covered all our cases. As discussed above, we excluded any state-based allegations. 

Section 5 and 12(a)(1) Violations of the Securities Act (``the Securities Act violations'', 15 U.S. Code § 77e and 15 U.S. Code § 77l \citep{noauthor_civil_nodate, noauthor_prohibitions_nodate}) were charged in both versions of the complaints for six of our cases, and Violations of Section 10(b) of the Exchange Act and Rule 10b-5 thereunder (``the Exchange Act violations'', 15 U.S. Code § 78j \citep{legal_information_institute_securities_nodate}) were included in both versions of the complaints for four cases. One case (\textit{Hong, et al. v. Block.One, et al.}) included both charges, so we asked one set of participants to evaluate it for the Securities Act violations and another for the Exchange Act violations. This, effectively, gave us a set of 10 cases (considering \textit{Hong, et al. v. Block.One, et al.} was included in the set twice). For full details of the charge selection process, see Appendix \ref{overlap}.



\subsubsection{Jury instructions}

We drafted jury instructions for each of our violations to determine the criteria upon which our mock jurors would decide if the relevant violations were proven in the complaint. We aligned the jury instructions for each of the violations we studied very closely with jury instructions for the same charges in real cases. 

We based our jury instructions for the Securities Act violations on \citep{dalton_sector_2013, ninth_circuit_jury_instructions_committee_manual_2023}. For the Exchange Act violations, we modeled our jury instructions on \citep{martin_plaintiff_2013}. The jury instructions each began with an introduction to securities laws and the claims in question. A series of five legal questions followed, each prefaced by relevant legal definitions and context. Each of the first four questions corresponds to one element of the charge that must be proven for the charge to be proven. The fifth question asked whether the juror answered the first four questions affirmatively (indicating a violation was proven). If they answered ``no'' to any of the questions, the allegation was not proven. In our study, we instructed the jury members to assume that the facts presented in the complaints were true. The legal questions were all binary (with ``yes'' or ``no'' answers). 

In a court case, the jury would need to decide on the legal elements of the case for each defendant individually. However, to simplify the process, because we are merely testing whether ChatGPT could write a convincing complaint based on certain facts, and because we abridged the amended complaints, we asked the jurors if they were satisfied that a particular element was supported—assuming all the facts alleged were true—for one or more defendants.

Following \cite{willmott_introduction_2018} and \cite{stevens_mock}, the final question of our jury instructions asked the jurors to rate their confidence in their final decision in the case on a scale of 1 to 5, with 5 meaning extremely confident and 1 meaning not at all confident.

Once we had drafts of our jury instructions, we piloted the survey for our sample complaint (which contained the Securities Act charges) on an individual without legal knowledge, noting the amount of time it took him to complete the survey and any confusion during the process. This led us to amend the instructions slightly: for example, they were edited to specify, in question four, that the plaintiff needed to have suffered \textit{financial} damages specifically. 

Table \ref{tab8} includes questions for each of our charges (excluding the background information). Our complete, final jury instructions can be found in Appendix \ref{juryinstructions1}. 

\begin{table}[htbp]
    \centering
    \begin{tabular}{p{5.5cm}|p{5.5cm}}
        \textbf{Securities Act} & \textbf{Exchange Act Violations} \\ \hline
       1. Assuming the facts alleged in the complaint you read are true, and noting the definitions of a security and sale thereof provided above, do you find that one or more of the defendants directly or indirectly sold securities to the plaintiff?  & 1. Assuming the facts alleged in the complaint you read are true, and noting the definitions provided above, do you find that one or more defendants (a) used a device, scheme or artifice to defraud, (b) made an untrue statement of material fact or made a statement that was misleading because a material fact was omitted, OR (c) engaged in any act, practice, or course of business which operated as a fraud or deceit upon any person? \\ \hline
       2. Assuming the facts alleged in the complaint you read are true, and noting the definitions of ``interstate commerce'' and ``instrument of transportation or communication'' provided above, do you find that one or more defendants used an instrument of transportation or communication in interstate commerce in connection with the offer or sale of a security? & 2. Assuming the facts alleged in the complaint you read are true, do you find that one or more defendants engaged in fraudulent conduct ``in connection with'' the purchase or sale of a security? \\ \hline
       3. Assuming the facts alleged in the complaint you read are true, do you find that ``the securities at issue weren't registered'' \citep{dalton_sector_2013}? & 3. Assuming the facts alleged in the complaint you read are true, do you find that one or more of the defendants acted knowingly or with severe recklessness? \\ \hline
       4. Do you find that the plaintiff suffered financial damages? & 4. Assuming the facts alleged in the complaint you read are true, do you find that one or more of the defendants' conduct involved interstate commerce, the use of the mails, or a national securities exchange? \\ \hline
       5. Did you answer ``yes'' to all of questions 1, 2, 3, and 4? & 5. Did you answer ``yes'' to all of questions 1, 2, 3, and 4? \\ \hline
       6. Thinking about your answers to questions 1-5, how confident are you, overall, that you have made the correct decision? Please rate your confidence on a scale of 1 to 5, with 5 being extremely confident and 1 being not confident at all. & 6. Thinking about your answers to questions 1-5, how confident are you, overall, that you have made the correct decision? Please rate your confidence on a scale of 1 to 5, with 5 being extremely confident and 1 being not confident at all. \\ \hline
    \end{tabular}
    \caption{Abridged jury instructions}
    \label{tab8}
\end{table}

\subsubsection{Survey}

We used Qualtrics to host our survey and Prolific to recruit our participants. Prolific provides pre-vetted candidates for research surveys \citep{noauthor_prolific_2014}. Each respondent received one of the 20 total complaints and the corresponding jury survey for the charge in that complaint. Our final data set included 88 respondents. For further details on our pre-screening, execution, demographic data, and quality control process, see Appendix \ref{prolific1}. 

\section{Results}

\subsection{GPT-3.5's ability to discern violations of federal U.S. law}

Overall, GPT-3.5’s ability to determine violations of U.S. federal law from a set of facts was poor, with an average final score across all cases and all runs ($n=100$) of 0.324 ($SD=0.317$) with a 95\% confidence interval around the mean of $[0.262 – 0.386]$. The full performance metrics are reported in Table \ref{tab2}. For five cases (\textit{Mark Winter, et al. v. Stronghold Digital Mining, Inc., et al.}, 1:22-CV-03088, SDNY;  \textit{Philip Golubowski, et al. v. Robinhood Markets, Inc., et al.}, 5: 21-CV-09767, N.D. Ca.;  \textit{Underwood, et al. v. Coinbase Global Inc.}, 1:21-cv-08353, SDNY; \textit{Securities and Exchange Commission v. Green United, LLC, et al.}, 2:23-cv-00159, D. Utah; \textit{Securities and Exchange Commission v. Paul A. Garcia}, 1:22-cv-00118, D. Colo.), the LLM did not determine any of the charges correctly in any of the outputs.

We note that, in general, precision (with a mean value of 0.66) tended to be higher than recall (which had a mean value of 0.25). This reflects the fact that the outputs were particularly prone to false negatives: when violations were identified, they tended to be correct, but the main shortcoming was the failure to identify violations. In \textit{Securities and Exchange Commission v. Arbitrade Ltd., et al.}, for example, GPT-3.5 produced false negatives between 8 and 8.5 of the 9 total charges, only identifying between 0.5 and 1 correct charge. This is more promising in terms of LLMs' usefulness as a tool to assist lawyers; it is much better for a case to exclude \textit{potentially} relevant allegations, than to include entirely incorrect ones.

\begin{table}[htpb]
    \centering
    \begin{tabular}{ c| c | c}
    \textbf{Metric} & \textbf{Mean} & \textbf{SD} \\ \hline
    Recall & 0.252 & 0.304 \\ \hline
    Precision & 0.658 & 0.459 \\ \hline
    Final score & 0.324 & 0.317 \\ \hline
    \end{tabular}
    \caption{GPT-3.5's ability to discern violations of U.S. securities laws, average metrics across all outputs ($n=100$)}
    \label{tab2}
\end{table}

\subsection{ChatGPT's pleading drafting ability}

The raw data from our surveys can be found in Appendix \ref{data}. Demographic data on the respondents can be found in Appendix \ref{prolific1}. 

\subsubsection{Respondent agreement}

To answer our question of whether jurors' decisions on complaints differed depending on their authors (AI versus a human lawyer), we first examined the levels of agreement among the reviewers for each of our cases. We examined this for the final decision as to whether or not the charge in question was proven (based on question 5, ``Did you answer ``yes'' to all of questions 1, 2, 3, and 4?''). The participants decided that the cases were proven in almost all instances ($n=74$), regardless of the author (84.1\%). Since we only sought to evaluate agreement based on this final (binary) question, we simply calculated the percentage of the total participants who agreed that the charge in question was proven.\footnote{\footnotesize{We acknowledge that there are various statistical measures used to calculate coder agreement, such as Cohen's Kappa. However, because the conclusion of each of our mock jurors was the answer to a single, binary question, using such a metric was not necessary in this case.}} Therefore, a percentage closer to either 0\% or 100\% (i.e., no jurors thought the charge was proven (0\%) or all the jurors agreed it was (100\%)) would indicate high levels of agreement, while a percentage closer to 50\% would indicate less agreement. As reported in Table \ref{tab3}, the level of agreement was moderate, overall, for most of our cases. The level of agreement was at chance (50\%) for three of our cases, two of which were drafted by ChatGPT and one of which was drafted by a lawyer. 

\begin{table}[htpb]
    \centering
    \begin{tabular}{p{7.5cm}| p{2cm} | p{1cm}}
         \textbf{Case} & \textbf{Number of Participants} & \textbf{\% Yes} \\ \hline
         \textit{Lee, et al. v. Binance, et al.}, GPT-drafted complaint	& 4	& 75\% \\ \hline
         \textit{Lee, et al. v. Binance, et al.}, lawyer-drafted complaint & 4	& 100\% \\ \hline
         \textit{Underwood, et al. v. Coinbase, et al.}, GPT-drafted complaint	& 6	& 83.3\% \\ \hline
         \textit{Underwood, et al. v. Coinbase, et al.}, lawyer-drafted complaint & 5	& 80\% \\ \hline
         \textit{Brola v. Nano, et al.}, GPT-drafted complaint	& 4	& 100\% \\ \hline
         \textit{Brola v. Nano, et al.}, lawyer-drafted complaint	& 4	& 75\% \\ \hline
         \textit{Ha v. Overstock.com, et al.}, GPT-drafted complaint & 4 & 75\% \\ \hline
         \textit{Ha v. Overstock.com, et al.}, lawyer-drafted complaint	& 5	& 100\% \\ \hline
         \textit{Hong, et al. v. Block.One, et al.}, GPT-drafted complaint, Securities Act charge	& 4	& 100\% \\ \hline
         \textit{Hong, et al. v. Block.One, et al.}, lawyer-drafted complaint, Securities Act Charge	& 4	& 100\% \\ \hline
         \textit{Hong, et al. v. Block.One, et al.}, GPT-drafted complaint, Exchange Act charge & 6	& 83.3\% \\ \hline
         \textit{Hong, et al. v. Block.One, et al.}, lawyer-drafted complaint, Exchange Act charge	& 4	& 50\% \\ \hline
         \textit{Balestra v. Cloud With Me Ltd.}, GPT-drafted complaint & 4 & 100\% \\ \hline
         \textit{Balestra v. Cloud With Me Ltd.}, lawyer-drafted complaint	& 4	& 100\% \\ \hline
         \textit{Audet, et al. v. Garza, et al.}, GPT-drafted complaint	& 5	& 50\% \\ \hline
         \textit{Audet, et al. v. Garza, et al.}, lawyer-drafted complaint	& 4	& 75\% \\ \hline
         \textit{Klingberg v. MGT Capital Investments Inc., et al.}, GPT-drafted complaint	& 4	& 50\% \\ \hline
         \textit{Klingberg v. MGT Capital Investments Inc., et al.}, lawyer-drafted complaint	& 4	& 100\% \\ \hline
         \textit{Davy v. Paragon Coin Inc., et al.}, GPT-drafted complaint	& 5	& 80\% \\ \hline
         \textit{Davy v. Paragon Coin Inc., et al.}, lawyer-drafted complaint	& 4	& 100\% \\ \hline
    \end{tabular}
    \caption{Respondent agreement on decisions}
    \label{tab3}
\end{table}

\subsubsection{Juror confidence}

We sought to examine whether juror confidence (on a scale of 1 to 5, with 1 being not confident at all and 5 being extremely confident) was different for complaints drafted by ChatGPT compared to those drafted by lawyers. To test this, we compared the mean confidences using a Mann Whitney U-test.\footnote{\footnotesize{We used this non-parametric test because our data were not normally distributed, as per the Shapiro-Wilk test, the results of which are reported in Appendix \ref{datadistribution}. We used the Mann-Whitney U test because our samples were still independent.}} The mean juror confidence for ChatGPT-generated complaints was 3.78, which a standard deviation of 0.99. For lawyer-drafted complaints, our jurors' mean confidence was 3.98, with a standard deviation of 0.84. There was no evidence that our jurors’ confidence scores differed between the two categories of authors ($U=1060$, $p=0.371$, $effect \ size = 0.095$ \citep{karadimitriou_mann-whitney_nodate}). This suggests that ChatGPT performed well at this legal drafting task.

\subsubsection{Association between author and juror decision}

Next, we looked at whether the author (human vs. ChatGPT) and the ultimate decision were associated in our cases using Fisher's Exact test. Table \ref{tab4} reports the two-way table for our data. 

\begin{table}[htpb]
    \centering
    \begin{tabular}{c| c| c | c | c}
        \multicolumn{2}{c}{} & \textbf{ChatGPT} & \textbf{Lawyer} & \textbf{Total} \\ \hline
        \textbf{Proven} & \textbf{No} & 9 (19.6\%) & 5 (11.9\%)	& 14 (15.9\%) \\ \hline
        & \textbf{Yes} & 37 (80.4\%) & 37 (88.1\%) & 74 (84.1\%) \\ \hline
        \textbf{Total} & & 46 & 42 & 88 \\ \hline
    \end{tabular}
    \caption{Two-way table for author and juror decision}
    \label{tab4}
\end{table}

Fisher's Exact test indicated that the author was not statistically significantly associated with the decision made by the respondents ($p=0.39$). This, again, suggests that ChatGPT's performance at this legal drafting task was as good as that of a lawyer. 

\subsubsection{Qualitative analysis of complaints}\label{qual}

We qualitatively compared the complaints ChatGPT produced with those written by lawyers. Overall, it was difficult to distinguish the complaints based on the author. The primary difference between the complaints drafted by each category of author was that the AI-generated versions were slightly more concise and less detailed. In general, we observed that the AI-drafted complaints were more straightforward, an effect we report linguistic analyses and quantitative findings on in Section \ref{quantconc}.

Somewhat notably, ChatGPT performed the worst at drafting captions for complaints, and this section needed to be regenerated more often than other sections (though, still, seldom, in only the \textit{Lee, et al. v. Binance, et al., Underwood, et al. v. Coinbase Global Inc.} and \textit{Hong, et al. v. Block.One, et al.} cases). For all sections, ChatGPT tended to include some sections of the complaint outside of that specified in the prompt (including, sometimes, an entire, very short complaint, similar to that in \citep{perlman_implications_2022}). However, it appeared to focus primarily on the section included in the prompt.

In one case, \textit{Hong, et al. v. Block.One, et al.}, we noted the different sections of the complaint generated were slightly inconsistent. For example, in the jurisdiction section, the complaint refers to violations of Section 12(a)(2) of the Securities Act, while the actual charge included in the legal claims for relief was Section 12(a)(1) thereof. Furthermore, in another case, \textit{Audet, et al. v. Garza, et al.}, the complaint includes different locations for one of the parties in different sections.

One of the key issues with using ChatGPT for legal tasks that others have noted as a risk (for example, \citep{choi_chatgpt_2023}), is its tendency to hallucinate or ``fabricate'' facts or statutes. We did notice this in two instances, namely that ChatGPT added ``John Doe'' defendants in two complaints (\textit{Hong, et al. v. Block.One, et al., Klingberg v. MGT Capital Investments Inc., et al.}). Our finding that ChatGPT hallucinated defendants in these two cases is consistent with our observation that ChatGPT was least successful in generating the output for the caption section.

We noticed that the complaint quality varied slightly among different jurisdictions (for example, the \textit{Klingberg v. MGT Capital Investments Inc., et al.} complaint in the District of New Jersey was of slightly poorer quality than the \textit{Underwood et al. v. Coinbase Global Inc.} one in the SDNY), though we did not quantify this effect. Furthermore, the ChatGPT-generated complaint for the \textit{Lee, et al. v. Binance, et al.} case was of slightly lower quality than some of the others. We hypothesized that this was because the article for the case was less focused (at least, as per our qualitative evaluation), and included facts from several other cases. We test this hypothesis in Section \ref{quantconc} below.

Finally, an important consideration in terms of ChatGPT's legal drafting ability is the level of human intervention required to produce the final complaint. In our study, post-editing the lawyer-drafted complaints to exclude facts that were not present in the Law360 article was more labor-intensive than post-editing the ChatGPT-generated complaints. Aside from the two cases where we needed to remove the additional ``John Doe'' defendants ChatGPT hallucinated, and where we needed to revise one reference to the specific section of the Securities Act that was violated, the remainder of the deletions occurred where ChatGPT had inserted placeholder text, the need for specific details should be inserted (such as the Plaintiff's residence or the dates of the class period), which were absent from the Law360 article. The only other time we needed to remove information ChatGPT included was in the \textit{Audet, et al. v. Garza, et al.} complaint, where we excluded the specific (and different) locations it generated for a party in different sections of the complaint.

Finally, we noticed a few stylistic differences between ChatGPT- and lawyer-drafted complaints. ChatGPT did not include any sub-headings in its sections, though these are quite common in lawyer-drafted complaints. Furthermore, ChatGPT frequently (and without any seeming purpose) added ``wherefore'' before sentences. 

\subsubsection{Quantifying complaint concreteness} \label{quantconc}

Our primary qualitative finding that the ChatGPT-drafted complaints were more straightforward than the lawyer-drafted ones motivated us to quantify this effect. We used the concept of concreteness, from the field of computational linguistics, to test this \citep{kleinberg_detecting_2019}. In language, concreteness means ``the degree to which the concept denoted by a word refers to a perceptible entity'' \citep{brysbaert_concreteness_2014}. We measure concreteness using \citeauthor{brysbaert_concreteness_2014}'s (\citeyear{brysbaert_concreteness_2014}) concreteness dictionary. In developing this data set, the researchers asked 4,000 human participants to score almost 40,000 words on a scale of 1 to 5, with 1 meaning abstract and 5 meaning concrete. Abstract words have ``meanings that cannot be experienced directly but which we know because the meanings can be defined by other words'', while concrete ones are those which one ``can experience directly through one of the five senses'' \citep{brysbaert_concreteness_2014}. 

To calculate the concreteness of our complaints, we tokenized the text from each complaint. For each token within each text, we obtained the average concreteness score therefor from the \cite{brysbaert_concreteness_2014} data set. We used code published by \cite{kleinberg_detecting_2019} for this task. The average concreteness score of all the tokens for each complaint constituted our final concreteness score. We used a t-test to determine whether the concreteness scores for the complaints written by ChatGPT were significantly different from those written by lawyers.\footnote{\footnotesize{The distribution for our two samples was normal ($Shapiro\text{-}Wilk = 0.901$, $p=0.256$ for GPT-generated complaints, and $Shapiro\text{-}Wilk = 0.850$, $p=0.074$ for lawyer-generated ones). The variances for both of our samples were the same.}} The average concreteness score for the ChatGPT-drafted complaints was 2.30 ($SD=0.06$). For lawyer-drafted complaints, the average concreteness was lower, at 2.25 ($SD=0.06$). The concreteness scores for each complaint are reported in Appendix \ref{concreteness}. 

According to our t-test, the average concreteness scores were significantly different between the two authorship categories, with a large effect size ($t(8)=3.25$, $p=0.012$, $Cohen's \ d=1.08, 99\% \ CI = [-0.02, 2.18]$). 

Next, we explored whether the level of concreteness in the Law360 article presented to ChatGPT was associated with the level of concreteness in the complaint ultimately drafted by ChatGPT. To do this, we calculated the concreteness score for the text of each Law360 article ($M=2.38, SD=0.09$) fed to ChatGPT and used the correlation between the Law360 article's concreteness and the resulting complaint's concreteness ($M=2.30, SD=0.06$) to measure this effect. The Pearson correlation coefficient between the Law360 article's concreteness and the concreteness of the AI-generated complaint was $r=0.78$ ($95\% \ CI = [0.24, 0.95]$, $p=0.01$). This shows a statistically significant positive correlation between the Law360 article's concreteness and the concreteness of the resulting ChatGPT-generated complaint. The higher the concreteness of the Law360 article, the higher the concreteness of the complaint generated therefrom.

\section{Discussion}

\subsection{GPT-3.5’s ability to discern violations of U.S. law}

This paper tests whether GPT-3.5 can correctly determine which federal U.S. laws are potentially being violated from a given fact pattern. Overall, GPT-3.5’s ability to determine potential violations of U.S. law from a given set of facts was poor. The primary reason for the low overall performance metrics was the high number of false negatives (i.e., violations it missed which were charged in the complaint). Because of this, the precision scores tended to be much higher than the other performance metrics. 

This shows more promise in terms of LLMs' ability to conduct legal reasoning tasks on behalf of lawyers, at least in a preliminary capacity. It would be preferable for fewer, but correct, charges to be identified, as opposed to the inclusion of erroneous charges, at least from a malpractice perspective. However, the trade-off of minimizing the number of false positives is that more manual analysis is, then, required to ``fill out'' the appropriate violations. Therefore, using an LLM to assist in legal reasoning of this nature would only be even somewhat useful at a preliminary stage. In an ideal world, an LLM would be most useful to lawyers as a ``sense check'' to make sure they were not missing any violations, but the observed tendency towards minimizing false positives means it would not be useful, at present, in this regard. Further to this, however, the charging decision-making process ordinarily goes through several iterations involving many lawyers. In this sense, the benchmark to which we compare GPT-3.5's performance may be unreasonably high. 

Despite GPT-3.5 missing several of the violations included in the complaint, its overall analysis of the facts for the charges it determined correctly was good (though we can only evaluate this qualitatively), using the legal reasoning framework provided. This is consistent with \citeauthor{yu_legal_2022}'s (\citeyear{yu_legal_2022}) findings that prompts including the IRAC legal reasoning framework produced good results. Finally, we expect these models' performance to improve as they incorporate more data.

Moreover, securities class action cases, in particular, are often dismissed. In a study of 652 such cases filed between 2006 and 2010, 32\% were dismissed with prejudice. After the first complaint was filed, of 533 cases completed before the study and filed during the study period, 25\% were dismissed with prejudice and 34\% were dismissed without prejudice \citep{klausner_when_nodate}. Therefore, it is possible that several of the charges GPT-3.5 failed to identify could have later been dismissed or not proven at trial---in this sense, such charges could in fact be considered to be true, rather than false, negatives. In that case, if GPT-3.5 only charged violations later confirmed by a judge or jury, its performance would be considered as good as that of a lawyer (notwithstanding the potential strategic reasons for including charges that are subsequently dismissed). We were unable to test this because the cases were ongoing, however, it would be a fruitful avenue for future studies to explore. 

We note that there could be other variables that account for GPT-3.5’s overall exclusion of many of the relevant violations in the complaint. Overall, it may simply be more conservative in estimating violations of law than the lawyers who drafted the complaint. This is something with which future research could experiment in terms of prompt engineering, for example, by instructing an LLM to ``include all charges for which there is any possibility they might be viable.'' Generally, GPT-3.5 cited far fewer violations of law, with many outputs only suggesting a single violation. This could also relate to lawyers’ occasional tendency to ``stack'' charges, often to inspire defendants to settle more readily. This means our ground truth is not necessarily objectively consistent either. This is consistent with the high number of false negatives we saw across our data set.

Furthermore, it could be the case that the facts sections from some of these complaints were written more clearly than others, or at least written in a way that is more conducive to helping the LLM identify relevant violations of law. 

At times, GPT-3.5 did not follow the instructions in the prompt. For example, despite asking the model to be specific about the sections of statutes which were violated, sometimes it did not even include the name of a statute at all, let alone the sections violated. Furthermore, the instructions requested it provide violations of federal law only; however, at times, it included violations of state law as well. This was somewhat surprising, particularly compared to other research (such as \citep{choi_chatgpt_2023}), which found it followed instructions well.

Aside from this, our results are generally consistent with previous research. For example, \citeauthor{blair-stanek_can_2023} (\citeyear{blair-stanek_can_2023}) also found that GPT-3’s statutory reasoning skills are lacking. Similarly, \citeauthor{choi_chatgpt_2023} (\citeyear{choi_chatgpt_2023}), noted an example of an ``issue spotter'' question (which refers to a similar task as in our study) in their test; ChatGPT only spotted one of the five correct issues correctly.

\subsection{ChatGPT’s pleading drafting ability}

This study also examined whether mock jurors made different legal decisions based on pleadings written by ChatGPT compared with those written by a lawyer. Overwhelmingly, ChatGPT drafted convincing complaints, which performed only slightly worse than the lawyer-drafted ones in terms of convincing jurors the allegations were proven (80\% of AI-drafted complaints were proven according to our jurors, compared to 88\% of lawyer-drafted ones). Our data, specifically our findings that complaint author was not associated with juror decision, and that the distribution of juror confidence did not differ between different authors, support these conclusions. 

We found the lawyer-generated complaints were less concrete than the AI-generated ones (though we note that, while statistically significant, the difference is minimal in absolute terms). This supports our qualitative findings that the AI-generated complaints were less abstract and more ``to the point'' than the lawyer-drafted ones. In practice, a lawyer using ChatGPT as a drafting assistant would add more abstraction (as per accepted legal drafting style) to the skeleton version the AI drafted. If lawyers increasingly use AI tools as drafting assistants, we could, however, see a future stylistic change in the industry, with less linguistic abstraction overall.

Notably, ChatGPT (and, indeed, lawyers), were able to write convincing complaints even for cases which were ultimately dismissed. In the case of the lawyer-drafted complaints, abridging these complaints (particularly removing specific charges) may have removed the sections upon which the judge’s ultimate dismissal decision hinged. This could explain, in part, why so many cases were proven according to our jurors. Furthermore, jurors do not reason the same way as judges, and, indeed, the criteria for dismissal are different than those required to prove a violation. Finally, the mock jurors in this study did not act like real-world jurors in that we instructed them to assume all facts in the complaint were true.

Our finding of a positive, statistically significant linear correlation between the level of concreteness in the Law360 article on which our ChatGPT-generated complaint was based, and the concreteness score of the resulting complaint, supports the conclusion that some of the complaint quality was likely associated with the level of detail in the Law360 articles which were provided as source material.

Other scholars have noted the possibility of ChatGPT having been trained on some legal sources \citep{bommarito_ii_gpt_2022}. We were concerned about this in answering our second research question, because we had to include cases which were filed before our model’s training period, due to a lack of cases filed only afterwards which met our selection criteria. However, consistently with \citeauthor{blair-stanek_can_2023}'s (\citeyear{blair-stanek_can_2023}) results, there was no indication (in our qualitative analysis of the complaints drafted) that made us suspicious that this was the case in our context (i.e., the ChatGPT-drafted complaint did not generally include facts which were not part of the article in the prompt). Hallucinations only occurred in three, minor instances. In two complaints (\textit{Hong, et al. v. Block.One, et al., Klingberg v. MGT Capital Investments Inc., et al.}), this took the form of ChatGPT inaccurately adding ``John Doe'' defendants, rather than devising specific, fictitious facts. In the third case (\textit{Audet, et al. v. Garza, et al.}), the location of a company involved was inconsistent across different sections of the complaint ChatGPT generated. This was a large, multinational company, and it may be the case that ChatGPT's training data included information about multiple of its locations (but, not likely about this case).

Our finding that complaints written for the SDNY were of a higher standard than those in the District of New Jersey, is intuitive, since more civil securities and commodity suits tend to be filed in the SDNY than any other jurisdiction, so the training data set likely would have included more cases from this jurisdiction, if it included any at all. For example, in March 2021, 36.8\% of suits of this nature were filed in the SDNY, compared to only 11.9\% for the next most popular filing jurisdiction \citep{trac_securities_2021}.

Overall, our results are consistent with prior research that qualitatively determined that ChatGPT showed some skill in drafting legal documents \citep{iu_chatgpt_2023, perlman_implications_2022}. Similarly, \cite{choi_chatgpt_2023}'s finding that ChatGPT performed better on essay questions than multiple choice ones is consistent with our finding that ChatGPT performed better in drafting than in statutory reasoning. 

\subsection{Limitations and future research}

Normally, jurors do not evaluate complaints, and, before a jury decides a case, they are presented with manifold evidence (and corroborating witness testimony) and summative arguments of the facts and legal issues from lawyers. They also have the opportunity to deliberate with their fellow jurors. In our study, we sought to obtain a baseline for juror decision-making based on a single document, assuming the facts, as presented, are true. Future research could explore conducting an entire mock trial using ChatGPT, rather than abstracting a specific element of the legal process as we have done, or employ lawyers to evaluate ChatGPT- and lawyer-generated complaints rather than laypeople.

Furthermore, there is no way to access all of the facts that the lawyers had when they were drafting the original complaints. We concluded that the best way to, therefore, ``feed'' ChatGPT the facts of each case would be to provide a summary of the facts and allegations from a reputable news outlet (in this case, we used Law360). One of the primary concerns with this is that the facts espoused in the article were pulled from the original complaint itself. However, since in this part of our research we are were merely testing ChatGPT's legal writing ability (i.e., its capacity to, given a set of facts and allegations, draft a pleading), we proceeded in this manner.

The lawyer-drafted complaints, also, naturally, frame the facts in a manner that supports the charges they assert. Assuming these framing choices are reflected in the Law360 article, it is possible that ChatGPT could produce convincing complaints by merely paraphrasing the article. Put simply, the input given to ChatGPT is already (indirectly) the product of some evaluation by a human lawyer, and so does not entirely reflect the ``raw'' form of the facts of the case. Again, since our aim is to obtain a baseline of ChatGPT's legal drafting capabilities, this input still allows this to be established; however, future research could explore performance when provided with more basic input. Finally, Law360 does require a subscription, which could hinder the reproducibility of results. However, they offer free trials which would be sufficient for this purpose. 

In the first part of our study, we decided not to test any cases that were filed before our model's training cut-off date in June 2021. Due to the lengthy nature of the legal process, these cases had generally not yet been decided. Therefore, we needed to use the charges in the complaints as our ground truth, though lawyers' judgments are not infallible---nor indeed are they always based purely on the factual ``correctness'' of charges. Lawyers may choose to file, or not to file, charges for other strategic reasons, and so the goals of lawyers may not be perfectly aligned with the task that an LLM is seeking to carry out. Future research could employ a set of lawyers to evaluate the LLM-generated charges, without considering strategic factors. Similarly, in the second part of our study, a judge's decision to deny or grant a motion to dismiss is subjective, and some could consider the decision incorrect. However, in this study, we sought to evaluate ChatGPT's drafting and decision-making, not that of lawyers and judges. 

Additionally, all our complaints required human intervention, whether drafted by ChatGPT or a lawyer. Overall, the post-edits for the lawyer-generated complaints were much more extensive than for the ChatGPT-generated ones. With a few, minor exceptions (discussed in detail in Section \ref{qual}), our amendments generally involved removing placeholder text ChatGPT provided. This level of intervention is consistent with our goal of evaluating ChatGPT's ability to assist lawyers in the legal drafting process. Future research could experiment with providing pure facts to ChatGPT and prompting the model to generate complaints in the same manner as we did---without any additional alterations---to determine if ChatGPT could perform satisfactorily as a lawyer on its own.

Finally, there are some limitations associated with using participants from Prolific in a study. First, some participants completed our survey very quickly ($median$ = 12:49:59 for all participants, including those whose entries were, ultimately, rejected), though we accounted for this in our quality control procedures, the details of which can be found in Appendix \ref{prolific1}. In any case, this kind of variability would be normal on a real-life jury as well. Furthermore, the high cost of using this platform meant we could not recruit more jury participants. This is somewhat exacerbated by the fact that few cases existed overall which met our inclusion criteria, making it slightly more difficult to generalise based on our results. 

Future research could use longer factual texts to determine whether the length affects the output, test these prompts on more powerful models, use different parameters to test these or related prompts, or test these prompt formats on other types of cases. Future research could also explore LLMs' legal reasoning and drafting capabilities in other jurisdictions. This may be particularly interesting in jurisdictions where case documents are not available online (and, therefore, they would be unlikely to have been trained on relevant examples thereof). 

\subsection{Conclusions: Do LLMs have the potential to meaningfully assist lawyers?}

In this study, we systematically evaluated GPT-3.5's and ChatGPT's potential usefulness in assisting with key elements of the legal process, namely legal reasoning and drafting. To this end, we used cryptocurrency securities cases as an example of the many potential areas of law where such support could be useful. Ours is the first study to systematically examine LLMs’ legal drafting and reasoning capabilities in a) litigation, b) the context of securities law, or c) in cases involving cryptocurrencies. We conducted two investigations—one involving asking GPT-3.5 an open-ended legal reasoning question, and the other a mock jury exercise requiring jurors to make decisions based on documents drafted by either ChatGPT or a human lawyer. Consistently with previous research, we found that the LLM was able to determine some federal civil laws which were potentially being violated in a given fact pattern from a real-world case; however, it tended to miss several salient violations. Nevertheless, we expect LLMs’ capabilities in this regard to improve as the models advance further. In contrast, again similarly to other preliminary empirical research and anecdotal evidence, ChatGPT performed well at the legal drafting task, namely drafting a complaint in a securities cryptocurrency class action lawsuit using facts from real-life cases. While GPT-3.5's current inability to perform a legal task as basic as legal reasoning means that it is unlikely to replace lawyers any time soon, ChatGPT's legal drafting skills may be able to support lawyers by reducing drafting time for key legal documents. 

\section*{Acknowledgements}

The authors thank Antonis Papasavva for participating in the pilot phase of our study and his valuable feedback. We also thank Anne Coventry for her assistance in conceptualizing our study and review of the legal concepts presented in this paper.

\section*{Competing interests}
The authors declare that they have no competing interests.

\section*{Funding}
This project was funded by UK EPSRC Grant EP/S022503/1 which supports the Centre for Doctoral Training in Cybersecurity at UCL.

\section*{Author's contributions}

All authors contributed to the study conception and design. Material preparation, data collection and analysis were performed by AT. The first draft of the manuscript was written by AT and all authors commented on previous versions of the manuscript. All authors read and approved the final manuscript.

\begin{appendices}
\section{Securities class action suit procedures}\label{civproced}

\subsection{U.S. civil procedure}\label{civproced1}

To commence a civil case in a U.S. federal court,\footnote{\footnotesize{Federal courts have jurisdiction over violations of federal law \citep{practical_law_federal_2023}. Federal Courts can also exercise diversity jurisdiction where the amount in controversy exceeds \$5 million and on the basis of geographic diversity between plaintiffs and defendants \citep{lender_class_2023}.}} the plaintiff (the party bringing the case) files what is called a complaint. This pleading is a legal document that includes the basis of the court's jurisdiction, the claims being brought, and what is referred to as a ``prayer for the relief'' that the plaintiff seeks in response to the defendant's alleged wrongdoing \citep{mcrae_initial_2023}. A complaint must include the following required elements:

\begin{itemize}
    \item A ``caption'' at the top, which is a heading that includes the court's name, the docket number (this is the case number, which the court assigns), the names of the parties, and ``Complaint'' as the document title. If the case is a class action matter (described below), this must also be included in the caption \citep{practical_law_litigation_commencing_2023}.
    \item A jury demand if the plaintiff wants a jury to decide the case \citep{practical_law_litigation_commencing_2023}.
    \item Various, specific sections in the body of the complaint:
        \begin{itemize}
            \item \textit{Preliminary statement}, which provides an overview of the case, ``identifies the main parties, important facts, legal causes of action, and relief sought'' \citep{practical_law_litigation_commencing_2023}. 
            \item \textit{Jurisdiction}, which briefly explains why the court has jurisdiction over this case.
            \item \textit{Venue}, which describes why this district court is the correct one to hear this case. District courts are federal trial courts in the U.S.
            \item \textit{Parties}, which describes the plaintiffs and the defendants in the case, including why the court has jurisdiction over the defendants and the subject matter in question.
            \item \textit{Facts}, which lays out the events and defendant's actions, on which the plaintiff's legal claims are based.
            \item \textit{Legal claims}, each of which is one ``Count'' against the defendant. Each count is a separate section and explains why the facts of the case support each allegation. At the end of each section, the plaintiff requests the appropriate relief from the court \citep{practical_law_litigation_commencing_2023}.
            \item \textit{Prayer for Relief}, ``reiterating the remedies sought for each count in the complaint'' \citep{practical_law_litigation_commencing_2023}.
            \item \textit{Signature block}, with the plaintiff's attorneys' information and a signature.
        \end{itemize}
\end{itemize}

All paragraphs in the complaint are numbered and there may be sub-headings within each section. Certain exhibits can be attached to the complaint to verify the allegations contained therein. ``If the plaintiff does not know all of the facts at the time it files the complaint, it may allege certain facts are true based on ``information and belief'''' \citep{practical_law_litigation_commencing_2023}.

Once filed with the court with the appropriate filing fee, the plaintiffs must serve this complaint on the defendant. The defendant must then respond to the complaint within a set period of time, either with an ``answer'' to the complaint, or a pre-answer motion, the most common of which is a motion to dismiss (MTD). An answer responds to the allegations in the complaint, offers general legal defenses to the allegations, and may include counterclaims against the plaintiff \citep{mcrae_initial_2023}. A MTD includes legal arguments as to why a case is not valid and should be thrown out by the judge. ``In ruling on a motion to dismiss, the court must accept the nonmoving party's allegations as true and usually may not consider extrinsic evidence'' \citep{mcrae_initial_2023}.

Next, the parties begin gathering facts about the case, preserve the necessary evidence, etc. \citep{mcrae_initial_2023}. What is called ``motion practice'' begins (these are called ``pre-trial'' motions if a trial is demanded by the plaintiff), whereby each party files various motions with the court (and legal memoranda or briefs that support these motions as well as relevant evidence) \citep{practical_law_litigation_motion_2023}. Motions are, essentially, lawyers' requests to the court for decisions on legal or procedural issues in connection with the case. There are various categories of motions. The first category is nondispositive motions, which include motions for other parties (not named in the case) to intervene, motions to amend pleadings, and motions for attorneys to be admitted \textit{pro hac vice} (meaning they wish to appear before a court in which they are not already formally admitted to practice) \citep{practical_law_litigation_nondispositive_2023}.

Discovery motions, which deal with evidence, are also filed. These include, for example, motions to compel discovery, quash or modify a subpoena, sanction parties for failure to preserve evidence, and other motions \citep{practical_law_litigation_discovery_2023}. Dispositive motions are also common, such as a motion for summary judgment, a MTD (if not already filed and decided), or a motion for default judgment \citep{practical_law_litigation_dispositive_2023}. 

Finally, attorneys may file evidentiary pre-trial motions, such as motions \textit{in limine}, which concern evidence to be presented to the jury, and \textit{Daubert} motions which relate to expert testimony \citep{weinberger_civil_2023}. The judge decides on all motions, sometimes at pre-trial hearings \citep{united_states_courts_civil_2022}.

Once pre-trial motions are all decided, the jury selection process begins for federal trials and jury instructions are submitted. Jury instructions include the legal standards for the case (i.e., the burden of proof), the laws that must be applied, how the jury should use the evidence presented at trial and what evidence they can consider, as well as other legal and practical information \citep{weinberger_civil_2023}. In general, the burden of proof in civil trials is the ``preponderance of evidence'' \citep{united_states_courts_civil_2022}. This means that the plaintiff must prove that it is more likely than not that violation of the law occurred.

During the trial, each side presents their opening statements, introduces evidence (through witnesses and exhibits), and presents a closing statement. Unless otherwise decided, the civil jury will contain at least six members and their decision must be unanimous \citep{weinberger_civil_2023}.\footnote{\footnotesize{If the jury does not reach a unanimous decision, the judge may instruct them to deliberate further or order a new trial \citep{noauthor_federal_2020}}.}

\subsection{Securities class action procedures under the PSLRA} \label{pslra}

A securities class action case begins when a plaintiff files a securities class action complaint. If similar actions have already been brought, the plaintiffs may decide to consolidate them. The court appoints a ``lead plaintiff'' for the class action. The defendant(s) then respond, usually with a MTD. Because of the requirements set forth in the PSLRA, many securities class actions do get dismissed. Cases may be dismissed for any of the following reasons:

\begin{itemize}
    \item Lack of jurisdiction of the court (either subject matter or personal);
    \item Improper process or service thereof;
    \item ``Improper venue'' \citep{practical_law_securities_litigaiton__white_collar_crime_securities_2023};
    \item ``Failure to state a claim'' (most commonly claimed in securities class action MTDs, as this reflects the underlying weakness of the case) \citep{practical_law_securities_litigaiton__white_collar_crime_securities_2023}; or
    \item ``Failure to join a party'' \citep{practical_law_securities_litigaiton__white_collar_crime_securities_2023}.
\end{itemize}

If the case survives dismissal, the next step is for the plaintiffs to request class certification by the judge. The general requirements for class certification are:

\begin{enumerate}
    \item ``Commonality, meaning there are questions of law or fact common to the class'' \citep{practical_law_securities_litigaiton__white_collar_crime_securities_2023};
    \item ``Typicality, meaning that the claims or defenses of the named plaintiff are typical of the class'' \citep{practical_law_securities_litigaiton__white_collar_crime_securities_2023}; and
    \item ``Adequacy, meaning that the named plaintiff must fairly and adequately protect the interests of the class'' \citep{practical_law_securities_litigaiton__white_collar_crime_securities_2023}. 
\end{enumerate}

The Federal Rules of Civil Procedure specify three categories of cases which can be brought as class actions. Every class action case must fit into one of these three categories. Securities class action suits generally fall under one of the following two categories:

\begin{enumerate}
    \item ``Common questions predominate over those affecting only individual members'' \citep{practical_law_securities_litigaiton__white_collar_crime_securities_2023}; or
    \item A class action lawsuit is the most efficient and fair way of resolving the case \citep{practical_law_securities_litigaiton__white_collar_crime_securities_2023}.
\end{enumerate}

The judge decides whether or not to certify the class and the discovery process ensues. In securities cases, in particular, the discovery process can be quite cumbersome due to the volume of documents involved and may last several years. This is one reason so many securities class actions settle. While the judge considers a MTD in a securities class action, discovery is automatically stayed; this is not true in a government enforcement case. At this stage, parties tend to file various motions, such as motions for summary judgment and \textit{Daubert} motions. The results of such motions tend to be leveraged in settlement discussions, which are ongoing throughout this process. If no settlement is reached, the case then proceeds to trial \citep{practical_law_securities_litigaiton__white_collar_crime_securities_2023}. 

\section{Prompt design process} 

\subsection{ChatGPT’s ability to discern violations of U.S. law} \label{prompt1}

The following link contains the prompts and outputs with which we experimented in developing the prompts to answer our first research question. It also describes our considerations as we fine-tuned the prompts: \url{https://osf.io/ejaxr?view_only=68c9693fa3994068b20d1be941b9db0b}. 

\subsection{Differences in juror decision-making based on human vs. AI-written legal complaints} \label{prompt2}

The following contains the prompts and outputs with which we experimented in developing the prompts to answer our second research question. It also describes our considerations as we fine-tuned the prompts: \url{https://osf.io/wszb2?view_only=68c9693fa3994068b20d1be941b9db0b}.

When we submitted our original prompts to the API, we received an error. We asked what the issue with this prompt was via the ChatGPT user interface, and it returned the following error message:

\begin{quote}
    ``The main issue with the prompt is that it is asking for the generation of a legal document, specifically the sections of a class action complaint for the SDNY, using the \textbf{\tt{openai.Completion.create()method}}. However, this method is designed for generating text based on a given prompt, and it is not suitable for generating complex legal documents with specific formatting and legal requirements.
    
    Generating a class action complaint requires specialized legal knowledge, attention to detail, and adherence to legal guidelines and formatting conventions. It is not appropriate to rely solely on an AI language model to generate such a document, as it may not be accurate, up-to-date, or tailored to the specific legal context. Legal documents should be prepared by qualified legal professionals who have expertise in the relevant jurisdiction and area of law.
    
    If you need assistance with legal matters, I recommend consulting with a licensed attorney who can provide you with the appropriate legal advice and assistance in drafting legal documents.''
\end{quote}

We received a similar error when we tested the same prompt on ChatGPT's user interface. We then added a disclaimer to the prompt, indicating the output would be ``for educational purposes only''. We still received an error using the API but were able to generate the desired text using the ChatGPT user interface, specifically the May 24, 2023 version. 

\section{Case selection}

\subsection{ChatGPT’s ability to discern violations of U.S. law} \label{cases1}

Details on our reasons for inclusion or exclusion for each cryptocurrency securities class action lawsuit and cryptocurrency-based SEC case filed before our model's training cutoff date (June 2021) can be found here: \url{https://osf.io/ecqas?view_only=68c9693fa3994068b20d1be941b9db0b}.

\subsection{Differences in juror decision-making based on human vs. AI-written legal complaints} \label{cases2}

Details on our reasons for inclusion or exclusion for each cryptocurrency securities class action lawsuit filed before our case selection date (May 22, 2023) can be found here: \url{https://osf.io/6we9r?view_only=68c9693fa3994068b20d1be941b9db0b}.

\section{Outputs and scoring of ChatGPT’s identification of potential violations of U.S. law} \label{results1}

The following contains the five GPT-3.5 outputs identifying potential violations of federal civil laws, including the number of input and completion tokens for each prompt: \url{https://osf.io/jgnea?view_only=68c9693fa3994068b20d1be941b9db0b}.


The following document includes our scores for each of the outputs found in \url{https://osf.io/jgnea?view_only=68c9693fa3994068b20d1be941b9db0b}, including the number of true positives GPT-3.5 identified (TP), false positives (FP), and false negatives (FN): \url{https://osf.io/txhy9?view_only=68c9693fa3994068b20d1be941b9db0b}.

The following contains the performance metrics for each of our outputs, calculated based on the scores awarded in \url{https://osf.io/txhy9?view_only=68c9693fa3994068b20d1be941b9db0b}: \url{https://osf.io/nhm48?view_only=68c9693fa3994068b20d1be941b9db0b}.

\section{Complaints} \label{complaints1}

The following links are to the various complaints presented to this study’s participants in answering our second research question:

\subsection{Sample case}

\begin{itemize}
    \item \textit{SEC v. Kaplan, et al.}, case number 2:23-cv-01648, in the U.S. District Court for the Eastern District of New York
        \begin{itemize}
            \item ChatGPT-drafted complaint: \\ \url{https://osf.io/gvfcn?view_only=68c9693fa3994068b20d1be941b9db0b}
            \item Abridged, lawyer-drafted complaint: \\ \url{https://osf.io/2gp8f?view_only=68c9693fa3994068b20d1be941b9db0b}
        \end{itemize}
\end{itemize}

\subsection{Dismissed cases}

\begin{itemize}
    \item \textit{Lee, et al. v. Binance, et al.}, case number 1:20-cv-02803, in the U.S. District Court for the Southern District of New York
        \begin{itemize}
            \item ChatGPT-drafted complaint: \\ \url{https://osf.io/8vmjs?view_only=68c9693fa3994068b20d1be941b9db0b}
            \item Abridged, lawyer-drafted complaint: \\ \url{https://osf.io/v796f?view_only=68c9693fa3994068b20d1be941b9db0b}
        \end{itemize}
    \item \textit{Underwood, et al. v. Coinbase Global Inc.}, case number 1:21-cv-08353, in the U.S. District Court for the Southern District of New York
     \begin{itemize}
            \item ChatGPT-drafted complaint: \\ \url{https://osf.io/8p5e3?view_only=68c9693fa3994068b20d1be941b9db0b}
            \item Abridged, lawyer-drafted complaint: \\ \url{https://osf.io/d4t9b?view_only=68c9693fa3994068b20d1be941b9db0b}
        \end{itemize}
    \item \textit{Brola v. Nano, et al.}, case number 1:18-cv-02049, in U.S. District Court for the Eastern District of New York
        \begin{itemize}
            \item ChatGPT-drafted complaint: \\ \url{https://osf.io/jxwu5?view_only=68c9693fa3994068b20d1be941b9db0b}
            \item Abridged, lawyer-drafted complaint: \\ \url{https://osf.io/zetrw?view_only=68c9693fa3994068b20d1be941b9db0b}
        \end{itemize}
    \item \textit{Ha v. Overstock.com, et al.}, case number 2:19-cv-00709, in the U.S. District Court for the District of Utah
        \begin{itemize}
            \item ChatGPT-drafted complaint: \\ \url{https://osf.io/572mq?view_only=68c9693fa3994068b20d1be941b9db0b}
            \item Abridged, lawyer-drafted complaint: \\ \url{https://osf.io/qwgzt?view_only=68c9693fa3994068b20d1be941b9db0b}
        \end{itemize}
\end{itemize}

\subsection{Continued cases}

\begin{itemize}
    \item \textit{Hong, et al. v. Block.One, et al.}, case number 1:20-cv-03829, in the U.S. District Court for the Southern District of New York
        \begin{itemize}
            \item ChatGPT-drafted complaint: \\ \url{https://osf.io/ze6cv?view_only=68c9693fa3994068b20d1be941b9db0b}
            \item Abridged, lawyer-drafted complaint: \\ \url{https://osf.io/7bxfh?view_only=68c9693fa3994068b20d1be941b9db0b}
        \end{itemize}
    \item \textit{Balestra v. Cloud With Me Ltd.}, case number 2:18-cv-00804, in the U.S. District Court for the Western District of Pennsylvania
         \begin{itemize}
            \item ChatGPT-drafted complaint: \\ \url{https://osf.io/yjqsw?view_only=68c9693fa3994068b20d1be941b9db0b}
            \item Abridged, lawyer-drafted complaint: \\ \url{https://osf.io/v6gu3?view_only=68c9693fa3994068b20d1be941b9db0b}
        \end{itemize}
    \item \textit{Audet, et al. v. Garza, et al.}, case number 3:16-cv-00940, in the U.S. District Court of Connecticut
        \begin{itemize}
            \item ChatGPT-drafted complaint: \\ \url{https://osf.io/7h2n8?view_only=68c9693fa3994068b20d1be941b9db0b}
            \item Abridged, lawyer-drafted complaint: \\ \url{https://osf.io/ptz69?view_only=68c9693fa3994068b20d1be941b9db0b}
        \end{itemize}
    \item \textit{Klingberg v. MGT Capital Investments Inc.}, et al., case number 2:18-cv-14380, in the U.S. District Court for the District of New Jersey
        \begin{itemize}
            \item ChatGPT-drafted complaint: \\ \url{https://osf.io/4tzju?view_only=68c9693fa3994068b20d1be941b9db0b}
            \item Abridged, lawyer-drafted complaint: \\ \url{https://osf.io/g3nqc?view_only=68c9693fa3994068b20d1be941b9db0b}
        \end{itemize}
    \item \textit{Davy v. Paragon Coin Inc., et al.}, case number 3:18-cv-00671, in U.S. District Court for the Northern District of California
        \begin{itemize}
            \item ChatGPT-drafted complaint: \\ \url{https://osf.io/tkm95?view_only=68c9693fa3994068b20d1be941b9db0b}
            \item Abridged, lawyer-drafted complaint: \\ \url{https://osf.io/8wq6m?view_only=68c9693fa3994068b20d1be941b9db0b}
        \end{itemize}
\end{itemize}

\section{Selecting overlapping charges} \label{overlap}

For each case, we listed the allegations present in the ChatGPT-generated complaint alongside the allegations from the original complaint. We then examined the overlapping charges between the two complaints for each case. One case (\textit{Aurelien Beranger, et al. v. Clifford ``T.I.'' Joseph Harris, et al.}, 1:18-cv-05054, N.D. Ga.) did not have any overlapping charges between the two complaints, so we excluded it at this stage.

Notably, in some cases (for example, \textit{Balestra v. Cloud With Me Ltd.}, in both the lawyer and the GPT version), only 12(a)(1) was charged but a finding of applicability of Section 5 is required for this charge.

A couple of cases included Section 20(a) control person liability violations, but these were less prevalent ($n=3$).  Because each of those cases also included at least one of the other charges we identified, we chose not to include this additional charge. 

In a couple of cases ($n=2$), the ChatGPT-drafted version of the complaint did not state the specific statutory violation, but rather a description thereof. For example, in one case, the relevant count was merely ``offering and sale of unregistered securities''. In this case, we inferred the violation of Sections 5(a), 5(c), and 12(a)(1) of the Securities Act.

\section{Jury instructions} \label{juryinstructions1}

In the following documents, we indicate where language was used from existing case documents (as is common practice). However, in providing these jury instructions to our mock jurors, we excluded this information to improve readability.

\subsection{Section 5 and Section 12(a)(1) violations}

The following link contains the jury instructions for cases alleging violations of the Securities Act: \url{https://osf.io/m9shg?view_only=68c9693fa3994068b20d1be941b9db0b}.

\subsection{Section 10(b) and Rule 10b-5 violations}

The following link contains the jury instructions for cases alleging violations of the Exchange Act: \url{https://osf.io/u2fn8?view_only=68c9693fa3994068b20d1be941b9db0b}.

\section{Survey execution and quality control} \label{prolific1}

We required all potential participants to have English as their first language, a minimum approval rate of 95\%, and to be U.S. citizens (since they would need to be eligible for jury duty). This gave us a set of 35,541 potential participants from Prolific’s pool of 120,471. 

We implemented a strict quality control process to determine our final participants, ultimately excluding 40 of the total 128 participants who completed our survey. Participants’ survey response times varied greatly, from 8 minutes and 51 seconds at the low end (of those accepted) to 2 hours and 30 minutes at the high end. We note that some of this variability can likely be explained by the differing lengths of our complaints, as well as participants’ level of familiarity with legal concepts. We suspect that those who complete mock jury studies frequently (which several participants indicated to us by message), would be able to complete our survey more quickly.

Based on our piloting process (of 4 participants), we expected someone who carefully read our study to take up to an hour and a half to complete it, and initially accounted for this amount of time in our budget. However, we soon noticed that many participants were much faster than this. We then adjusted the expected time of our survey slightly.

For participants who finished the survey in under 20 minutes, we manually calculated the amount of time it would take to read the survey in its entirety at various reading speeds. Based on the results of \cite{brysbaert_how_2019}, we used 238 wpm as our ``average'' reading speed, and 500 wpm as a very fast reading speed. We excluded anyone who was completing the survey in less time than it would take to read it at between 400 and 500 wpm.

Finally, we excluded any participants who sent us messages from whose messages it was clear that English was not their first language ($n=2$).

\subsection{Participant's demographic data}

In Table \ref{demo} we report the demographic data for the participants whose responses we approved and included in the results of this study (and who consented to the sharing of this information and for whom said data had not expired). 

\begin{table}[h]
    \centering
    \begin{tabular}{l|l}
         \textbf{Demographic marker} & \textbf{Makeup of approved participants} \\ \hline
         Age & \makecell[l]{Participants aged 23-73 \\ 15\% Under 30 \\ 38\% Aged 30-39 \\ 20\% Aged 40-49 \\ 13\% Aged 50-59 \\ 9\% Aged 60-69 \\ 5\% Aged 70-73}  \\ \hline
         Sex & \makecell[l]{37\% Female \\ 63\% Male}\\ \hline
         Ethnicity & \makecell[l]{1\% Asian \\ 13\% Black \\ 9\% Mixed race \\ 5\% Other \\ 73\% White}\\ \hline
         Country of birth & \makecell[l]{1\% Ghana \\ 1\% New Zealand \\ 1\% Nigeria \\ 1\% South Africa \\ 95\% United States of America} \\ \hline
         Nationality & 100\% United States of America \\ \hline
         Student status & \makecell[l]{11\% Yes \\ 89\% No}\\ \hline
         Employment status & \makecell[l]{56\% Full-time \\ 20\% Not in paid employment (e.g. homemaker, retired, or disabled) \\ 11\% Part-time \\ 6\% Unemployed (and job-seeking) \\ 7\% Other }\\ \hline
    \end{tabular}
    \caption{Study participants' demographic data}
    \label{demo}
\end{table}

\begin{figure}[htpb]
    \centering
    \includegraphics[width=12cm, keepaspectratio]{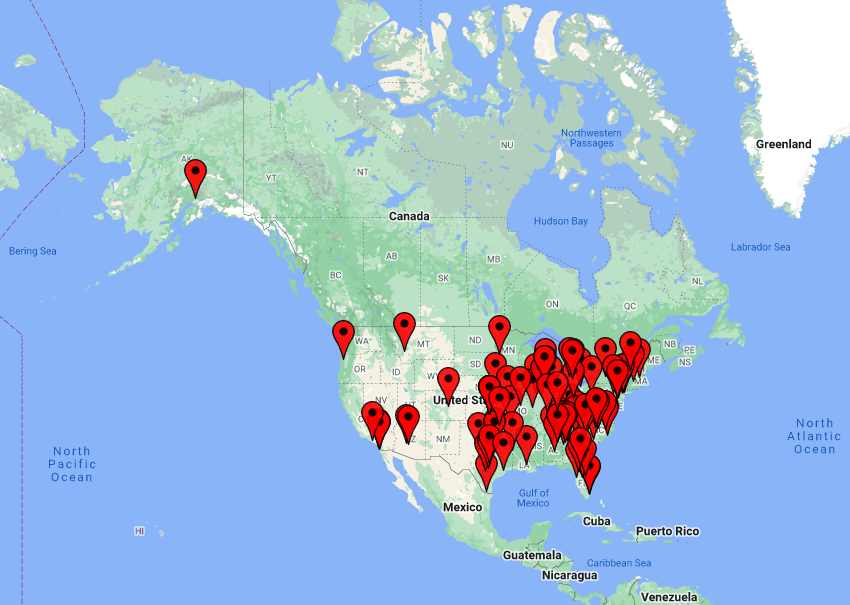}
    \caption{Accepted participants' locations}
    \label{location}
\end{figure}

In Figure \ref{location} we also include a map of our accepted participants' locations.

\section{Survey Data} \label{data}

The data from our Prolific survey can be found here: \url{https://osf.io/7xnmj?view_only=68c9693fa3994068b20d1be941b9db0b}.

\section{Data distribution} \label{datadistribution}

Table \ref{tab7} reports the results of Shapiro-Wilk tests on our data, showing that the data were not normally distributed.

\begin{table}[htpb]
    \centering
    \begin{tabular}{c|c| c}
         & \multicolumn{2}{c}{\textbf{Shapiro-Wilk (\textit{p}})} \\ \hline 
         & ChatGPT-authored & Lawyer-authored \\ \hline
         All cases & 0.771 ($<$ 0.001) & 0.772 ($<$ 0.001) \\ \hline 
    \end{tabular}
    \caption{Data distribution: juror confidence scores}
    \label{tab7}
\end{table}

\section{Concreteness scores} \label{concreteness}

\begin{table}[htpb]
    \centering
    \begin{tabular}{l|c}
        \textbf{Case} & \textbf{Concreteness} \\ \hline
         \textit{Lee, et al. v. Binance, et al.} & 2.28 \\ \hline
         \textit{Hong, et al. v. Block.One, et al.} & 2.26 \\ \hline
         \textit{Balestra v. Cloud With Me Ltd.} & 2.43 \\ \hline
         \textit{Underwood, et al. v. Coinbase, et al.} & 2.32 \\ \hline
         \textit{Audet, et al. v. Garza, et al.} & 2.25 \\ \hline
         \textit{Klingberg v. MGT Capital Investments Inc., et al.} & 2.33 \\ \hline
         \textit{Ha v. Overstock.com, et al.} & 2.23 \\ \hline
         \textit{Brola v. Nano, et al.} & 2.35 \\ \hline
         \textit{Davy v. Paragon Coin Inc., et al.} & 2.25 \\ \hline 
    \end{tabular}
    \caption{Concreteness scores for ChatGPT-generated complaints}
    \label{conc-tab-gpt}
\end{table}

\begin{table}[htpb]
    \centering
    \begin{tabular}{l|c}
         \textbf{Case} & \textbf{Concreteness} \\ \hline
         \textit{Lee, et al. v. Binance, et al.} & 2.20 \\ \hline
         \textit{Hong, et al. v. Block.One, et al.} & 2.22 \\ \hline
         \textit{Balestra v. Cloud With Me Ltd.} & 2.32 \\ \hline
         \textit{Underwood, et al. v. Coinbase, et al.} & 2.22 \\ \hline
         \textit{Audet, et al. v. Garza, et al.} & 2.24 \\ \hline
         \textit{Klingberg v. MGT Capital Investments Inc., et al.} & 2.28 \\ \hline
         \textit{Ha v. Overstock.com, et al.} & 2.23 \\ \hline
         \textit{Brola v. Nano, et al.} & 2.36 \\ \hline
         \textit{Davy v. Paragon Coin Inc., et al.} & 2.20 \\ \hline 
    \end{tabular}
    \caption{Concreteness scores for lawyer-generated complaints}
    \label{conc-tab-law}
\end{table}

\begin{table}[htpb]
    \centering
    \begin{tabular}{l|c}
         \textbf{Case} & \textbf{Concreteness} \\ \hline
         \textit{Lee, et al. v. Binance, et al.} & 2.29 \\ \hline
         \textit{Hong, et al. v. Block.One, et al.} & 2.35 \\ \hline
         \textit{Balestra v. Cloud With Me Ltd.} & 2.49 \\ \hline
         \textit{Underwood, et al. v. Coinbase, et al.} & 2.34 \\ \hline
         \textit{Audet, et al. v. Garza, et al.} & 2.32 \\ \hline
         \textit{Klingberg v. MGT Capital Investments Inc., et al.} & 2.49 \\ \hline
         \textit{Ha v. Overstock.com, et al.} & 2.25 \\ \hline
         \textit{Brola v. Nano, et al.} & 2.51 \\ \hline
         \textit{Davy v. Paragon Coin Inc., et al.} & 2.39 \\ \hline
    \end{tabular}
    \caption{Concreteness scores for Law360 articles}
    \label{conc-tab-law360}
\end{table}




\end{appendices}

\clearpage

\bibliography{sn-biliography}

\end{document}